\documentclass{article}
\usepackage{PRIMEarxiv}
\usepackage{amsthm,amsmath,amsfonts,latexsym,amssymb}
\usepackage{graphics,fancyhdr,graphicx}
\usepackage[ruled,linesnumbered]{algorithm2e}
\usepackage{textgreek}
\usepackage{listings}
\usepackage[utf8]{inputenc} 
\usepackage[T1]{fontenc}    
\usepackage{hyperref}       
\usepackage{url}            
\usepackage{multirow,booktabs}       
\usepackage{amsfonts}       
\usepackage{nicefrac}       
\usepackage{microtype}      
\usepackage{lipsum}
\usepackage{fancyhdr}       
\usepackage{graphicx}       
\usepackage[font={small,it}]{caption}
\usepackage{listings}
\usepackage[justification=centering, font=small]{subcaption}
\usepackage{placeins}
\usepackage{comment}
\usepackage{lscape}
\usepackage{float}
\usepackage{soul}
\usepackage{tabularx}
\usepackage{amssymb}
\usepackage{lineno}
\usepackage{comment}
\usepackage{bm}
\usepackage{tcolorbox}
\usepackage{cite}
\graphicspath{{media/}}     

\title{MyCrunchGPT: A chatGPT assisted framework for scientific machine learning
}

\author{Varun Kumar \thanks{School of Engineering,
  Brown University,
  Providence, RI.}\\
  \texttt{varun\_kumar2@brown.edu}
  \And
   Leonard Gleyzer \thanks{ Division of Applied Mathematics,
  Brown University,
  Providence, RI.}\\
  \texttt{leonard\_gleyzer@brown.edu} \\
  \And
   Adar Kahana \footnotemark[2] \\
  \texttt{adar\_kahana@brown.edu} \\
  \And
   Khemraj Shukla \footnotemark[2] \\
  \texttt{khemraj\_shukla@brown.edu}\\
  \And
  George Em Karniadakis
  \footnotemark[1] \; \footnotemark[2] \; \thanks{Corresponding author}\\
\texttt{george\_karniadakis@brown.edu} \\
}

\begin{document}
\maketitle

\begin{abstract} 
Scientific Machine Learning (SciML) has advanced recently across many different areas in computational science and engineering. The objective is to integrate data and physics seamlessly without the need of employing elaborate and computationally taxing data assimilation schemes. However, preprocessing, problem formulation, code generation, postprocessing and analysis are still time-consuming and may prevent SciML from  wide applicability in industrial applications and in digital twin frameworks. Here, we integrate the various stages of SciML under the umbrella of ChatGPT,
to formulate {\em MyCrunchGPT}, which plays the role of a conductor orchestrating the entire workflow of SciML based on simple prompts by the user. Specifically, we present two examples that demonstrate the potential  use of MyCrunchGPT in optimizing airfoils in aerodynamics, and in obtaining flow fields in various geometries in interactive mode, with emphasis on the validation stage. To demonstrate the flow of the MyCrunchGPT, and create an infrastructure that can facilitate a broader vision, we built a webapp based guided user interface, that includes options for a comprehensive summary report.
The overall objective is to extend MyCrunchGPT to handle diverse problems in computational mechanics, design, optimization and controls, and general scientific computing tasks involved in SciML, hence using it as a research assistant tool but also as an educational tool. While here the examples focus in fluid mechanics, future versions will target solid mechanics and materials science, geophysics, systems biology and bioinformatics.
\end{abstract}

\keywords{ChatGPT for design \and NACA airfoil optimization \and ChatGPT for PINNs \and fluid mechanics \and cavity flow \and validation}

\section{Introduction}
With the advancements in deep learning and natural language processing techniques in recent times, Large Language Models (LLMs) have witnessed remarkable development and research interest, both from academia and industry alike. The availability of large text corpus in the form of digital content, coupled with modern language processing techniques such as Transformers \cite{vaswani2017attention}, has improved the efficacy and accuracy of natural language processing tasks such as language translation, text classification and sentence completion. Some state-of-the-art LLMs include BERT \cite{devlin2018bert}, T5 \cite{raffel2020exploring}, RoBERTa \cite{liu2019roberta}, XLNet \cite{yang2019xlnet} amongst others. These models are typically trained on large text corpus, contain billions of neural network parameters, and can be adapted to meet a user's specific use case using these pre-trained models \cite{kumar2020data, shah2021gt, rothe2020leveraging}. A significant and potentially breakthrough advancement was achieved by OpenAI through the development of Generative Pre-trained Transformer (GPT-3), which has shown remarkable performance for a range of natural language processing tasks ranging from sentence completion to providing specific responses based on instructions provided by the user \cite{brown2020language}. Colloquially named as ChatGPT, the GPT language model is one of the largest natural language processing models (with hundreds of billions of parameters)  that exists today and is capable of generating very high-quality, human-like responses to user queries on a wide range of topics.

A key ability of ChatGPT model that makes it a significant improvement over existing language models is its ability to learn and comprehend instructions provided by the user and respond appropriately. In other words, a user can define a desired `persona' for ChatGPT, whereby the responses generated by ChatGPT to user queries can be made akin to human-like response. This opens up the possibility of using ChatGPT for process automation in engineering and natural sciences \cite{blanco2022role, bran2023chemcrow, frieder2023mathematical, leiter2023chatgpt, maddigan2023chat2vis, prieto2023investigating, hassani2023role, bishop2023computer}. With its ability to generate suitable code or responses for repetitive, labour-intense tasks, which are often encountered in engineering and natural sciences alike (for instance, writing codes for visualization or creating a patient after-visit summary), there exists an opportunity to use the generative capability of ChatGPT for improving processes with well-defined goals. This idea of using the capability of AI tools to automate decision-making has been explored in connection with expert systems since 1970s. Dendral and Meta-Dendral \cite{buchanan1978dendral} developed pioneering work that combined the capability of machine learning to identify molecular structures of unknown compounds. This work helped demonstrate the capability of AI-powered expert systems to automate domain-specific complex tasks. 

In this present work, we develop a ChatGPT-based assistant, called MyCrunchGPT (or simply MyCrunchGPT), to streamline two specific tasks: (1) design and optimization of 2D NACA airfoils, and (2) solving prototype fluid mechanics problem using PINNs \cite{raissiPINNs}. The design process, as we discuss later, involves a series of steps starting from concept generation for detailed design. We focus on the conceptual design stage to demonstrate the application of MyCrunchGPT Design Assistant that provides a natural-language based interactive environment to guide the user through design and the specific optimization process of a 2D NACA airfoil. The Design Assistant helps automate the specific task to generate new airfoil designs, analyze these new designs using surrogate models, help the user to optimize the airfoil shape and visualize the results when requested by the user. More broadly, we aim to leverage the idea of generative design \cite{buonamici2020generative, kalliorasgenerative, krishgenerative, gruber2020generative, shea2005generative}, a process that involves exploration of larger design spaces using machine learning algorithms, to augment a designer's options during the conceptual design stage.  The key idea here is to use DeepOnet \cite{lu2021learning}, the first deep neural operator that can play the role of a surrogate that can generalize to different airfoil geometries. 
We also demonstrate how physics-informed neural networks (PINNS) \cite{raissiPINNs}  can be used in MyCrunchGPT to compute and analyze the results of prototype flows, and how the user by initiating simple prompts can change aspects such as the problem geometry and training hyperparameters so that the underlying code implementation can change accordingly. The same prompting process can be used in solid mechanics and other simulation and analysis tasks in computational science and engineering. 

\subsection{Objective and contributions} 
The primary objective of this work is to set the foundations for an integrated framework that uses LLMs to simplify use of SciML in everyday tasks in computational science and engineering, accelerating engineering design and scientific discovery. While the conceptual framework is general, the currently implemented framework was dictated by the logistic constraints and lack of flexible APIs for the current generation GPT-4. Within this context, the specific contributions of our work are:

\begin{itemize}
\item Develop MyCrunchGPT as an assistant to guide users through specialized processes such as engineering design. As an example, we demonstrate 2D NACA airfoil design, optimization, and verification using a standard CFD solver.
\item Demonstrate the use of MyCrunchGPT for assisting users in solving classical fluid mechanics problems using PINNs in various geometries.
\item Develop a generic framework for integrating MyCrunchGPT with specialized workflow via a custom interactive web-based interface.
\item Facilitate an easy human-computer interaction with a web-based framework and generate an end-to-end report of interaction along with an executive summary.  
\end{itemize}

\section{Scientific Machine Learning (SciML)} 
\label{sec:headings}
SciML has emerged as a new form of scientific computing that has the potential to  affect greatly all areas of computational science and engineering. It is based on the realization that neural networks can be universal approximations of functions but also of differential operators, making it possible to solve complex systems of partial differential equations (PDEs) or parametrized PDEs while seamlessly injecting any scattered/noisy spatiotemporal data into the solution. This new form of non-sterilized computing has been quickly adopted by the industry for diverse applications \cite{NVIDIA_modulus,bosch_pinn}. However, many challenges remain associated with neural architecture search, hyperparameter tuning, solution verification and uncertainty quantification. In addition, there is the added cost associated with the problem formulation, data preprocesing, code generation, visualization and analysis, and uncertainty quantification. We propose to overcome all these barriers by building an integrated end-to-end framework that automates all these processes using chatGPT as the main front-end and in-house methods such as PINNs for PDEs, neural operators such as DeepONet surrogates and for real-time time forecasting \cite{kumar2023real}, but also for other uses, e.g., constructing fast solvers in scientific computing, e.g. see HINTS, \cite{zhang2022hybrid}. Taken together these developments have the potential to make MyCrunchGPT a friendly and effective research assistant that can continually learn by incorporating new data, new methods, and new LLMs in a seamless fashion. Of particular interest is to produce solutions, which can be verified and validated so that they satisfy the physical principles, e.g. conservation laws, and agree with available experimental data and existing theories that can be readily provided by chatGPT.

\subsection{Physics-Informed Neural Networks} 
Physics-Informed Neural Networks (PINNs) were first introduced in \cite{PINNs}, providing a new paradigm for obtaining numerical solutions to both forward and inverse PDEs. PINNs work by representing the PDE solution as a neural network and learning the solution in a semi-supervised learning fashion by data provided by measurements and the PDE evaluated at random points in the spatiotemporal domain. Automatic differentiation is leveraged to encode the PDE residual of the predicted solution into the loss function. A schematic of the structure of a PINN is given in Figure 1.\\

\begin{figure}[h]
    \centering
    \includegraphics[scale=0.35]{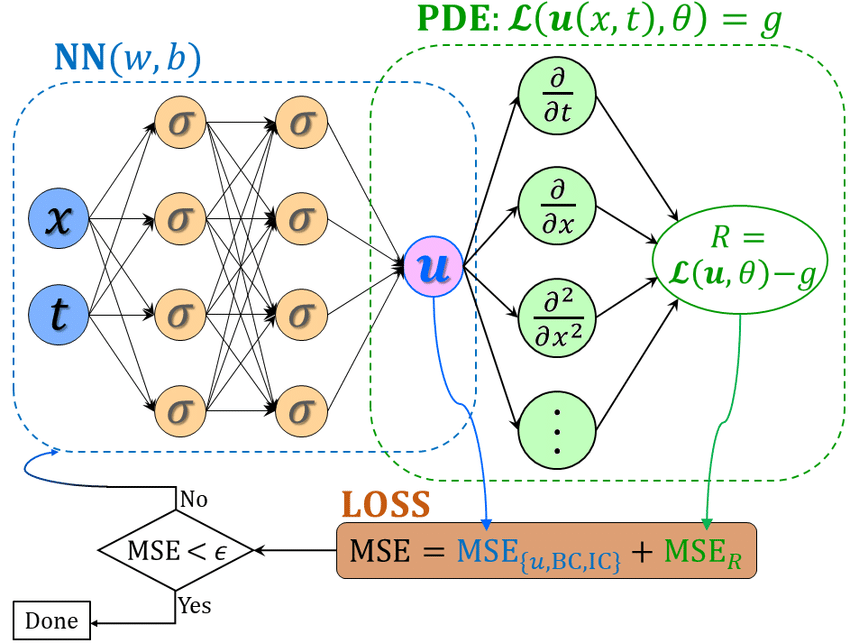}
    \caption{Schematic of a vanilla Physics-Informed Neural Network, figure from \cite{Meng_2020}. The top-left component (NN) represents the neural network solution $u_\textit{NN}$ of the unknown function $u$. In the top-right component (PDE), automatic differentiation is used to get the necessary derivatives of the output $u$ w.r.t. its inputs $(x, t)$. Those derivatives are then combined to compute the residual $R$ between the PDE operator $\mathcal{L}$ applied to the learned solution $u_\textit{NN}$ and the known forcing function $g$. The loss is then a combination of the standard supervised MSE loss applied to the known boundary/initial conditions (and possibly known values at points within the domain), and the MSE of the residual $R$.}
    \label{fig:pinn_schematic}
\end{figure}

Since their inception, many variants of PINNs have been developed, including (but not limited to) conservative, Bayesian \cite{Yang_2021}, fractional \cite{Pang_2019}, parareal \cite{Meng_2020}, variational \cite{kharazmi2019variational}, stochastic \cite{yang2018physicsinformed}, among others. Since the PINN framework is fairly generic, it lends itself well to problems across a variety of domains. Examples of problems in which PINNs have demonstrated promising results include: Predicting the severity of brain aneurysms using blood flow analysis from MRI videos \cite{cai2023aiv}; Modeling and predicting the behavior of the COVID Omicron variant \cite{Cai_2022}; Inferring 3D velocity and pressure fields from tomographic background oriented Schlieren videos \cite{Cai_2021}; quantifying microstructure material properties from ultrasound data \cite{Shukla_2022}; Nonhomogeneous material identification in elasticity imaging \cite{zhang2020physicsinformed}; Predicting gas flow dynamics and unknown parameters in diesel engines \cite{nath2023physicsinformed}.

\subsection{Deep Operator Network (DeepONet)} 
A vanilla DeepONet \cite{lu2021learning} consists of two deep neural networks, the branch and the trunk net. The branch net encodes the input function $\boldsymbol u$ at $m$ fixed locations, also known as sensor locations. The branch network can comprise of multi-layer fully connected network/s or other conventional deep learning architectures such as CNN, RNN, and LSTM. The input function $\bm{u}$ to the branch network is discretized at $m$ sensor locations such that $\bm{u}^{(i)} = \{u^{(i)}(\bm x_1), u^{(i)}(\bm x_2), \ldots, u^{(i)}(\bm x_{m})\}$ and $i \in [1,n]$, where $n$ is the number of samples. The coordinate locations, $\bm{y} = \{{\bm y_1, \bm y_2, \ldots, \bm y_{q}}\}$ are added as trunk network input to generate the basis function where the solution operator will be evaluated. The solution operator,  $\mathcal G_{\bm \theta}$ is constructed as an element wise product of the output embeddings of the branch and the trunk networks. The optimized network parameters, $\boldsymbol {\theta}^*$ are obtained by minimizing a loss function ($\mathcal L_1$ or $\mathcal L_2$), defined as
\begin{equation} \label{eq:L1_loss}
\begin{split}
    \mathcal L_1 &= \frac{1}{n}\sum_{i =1}^n \sum_{j =1}^q \big| \mathcal G(\bm u^{(i)})(\bm y_j) - \mathcal G_{\bm{\theta}}(\bm u^{(i)})(\bm y_j)\big|\\
    \mathcal L_2 &= \frac{1}{n}\sum_{i =1}^n \sum_{j =1}^q\big(\mathcal G(\bm u^{(i)})(\bm y_j) - \mathcal G_{\bm{\theta}}(\bm u^{(i)})(\bm y_j)\big)^2,\\
\end{split}
\end{equation}
where $\mathcal{G}_{\bm{\theta}}(\bm u^{(i)})(\bm y_j)$ is the predicted value obtained from the DeepONet, and $\mathcal G(\bm u^{(i)})(\bm y_j)$ is the target value. For more details on DeepONet formulation, readers can refer to \cite{goswami2022physics}.

\begin{figure}[h]
  \centering
  \centerline{\includegraphics[width = 0.9\textwidth]{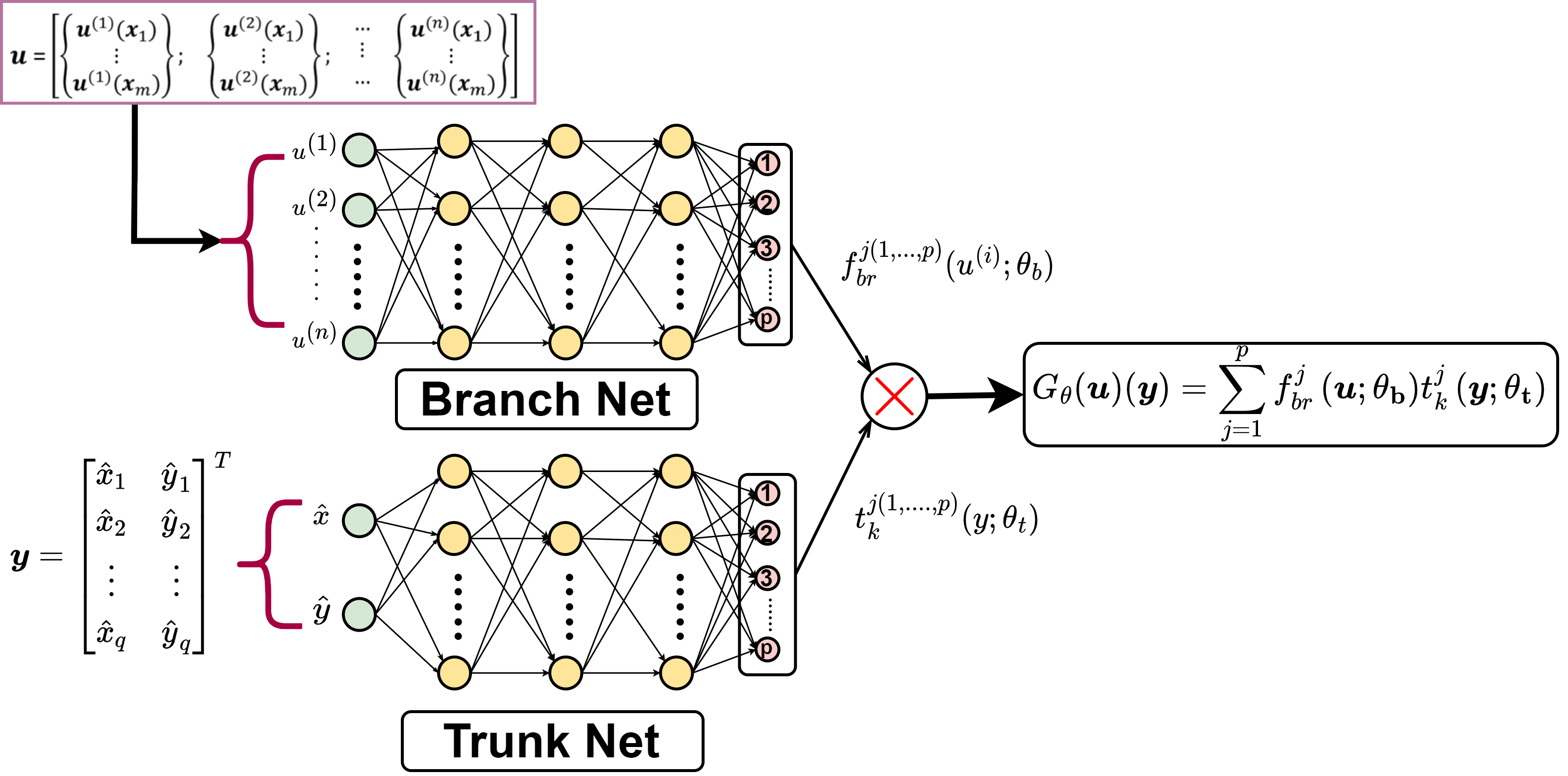}}
  \caption{Vanilla DeepONet architecture showing branch and trunk networks (modified from \cite{goswami2022physics}). The branch network takes functions $u^{i (1...n)}$ as inputs evaluated at $m$ sensor locations while the trunk network takes continuous coordinates and parameters, $\bm{y}$ as inputs.}
  \label{fig:vanilla_don}
\end{figure}

\section{MyCrunchGPT} 
MyCrunchGPT uses chatGPT as the front-end and various in-house programs at the back-end that implement PINNs and DeepOnet, supplemented by various modules associated with the geometry, visualization and analysis, etc. These modules and capabilities will be expanded in the near future as GPT-4 APIs become available but for now we present how we use DeepOnet and PINNs in MyCrunchGPT.

\subsection{MyCrunchGPT for Operator-based Design}
A typical design process starts with a statement for a product, which is transformed into a set of technical requirements. Engineering design can be described as a systematic process for solving a given problem under the stated constraints. The engineering design process generally involves three stages: conceptual design, embodiment design, and detailed design \cite{beitz1996engineering}. The conceptual design stage typically involves idea generation, design space exploration, and preliminary design evaluation. The methodology adopted for conceptual design varies based on the application setting, sometimes even based on user's personal preferences. Over time, designers develop specialized knowledge base and design methodology based on experience. This presents an opportunity to capture this knowledge-base and experience in an intelligent framework that can be shared and used concurrently by other designers.

Preliminary analysis of design concepts generated during this stage is essential to enable concept selection for further refinement. In engineering design, the analysis tools include CAE, CFD, lumped-system models, and other model-based tools. Often, these tools are computationally expensive and time consuming, especially when multiple designs need to be analyzed. To augment the analysis phase, pre-trained Operator-based Deep Neural networks are useful tools, with their ability to generate analysis results within seconds instead of hours. The DeepONet is generally trained for a specific task, and can help replace or augment the computationally expensive tools that can sometimes inhibit analysis during concept design phase. The ability to analyze multiple designs rapidly also adds value to the design process, since design is typically an iterative process. Hence, an operator-based network such as the DeepONet is an integral part of our framework and acts as a fast surrogate for analyses.

MyCrunchGPT for Deep Operator-based Design presents a structured approach to conceptual design that uses an AI design assistant to guide the user through the conceptual design generation process. The design assistant provides a high level interface to guide the user through a step-by-step approach to accomplish a specific design task. The design assistant uses chatGPT's ability to generate domain-specific responses to user's queries. With chatGPT's ability to understand and generate context-based response, the design assistant is tailored to meet specific use cases that are typically encountered during the design process. Additionally, chatGPT being a LLM, the response generated by the design assistant can also be modified based on user's needs. For instance, the design assistant can enable a novice design engineer to explore specific subjects while guiding through the design process. The assistant's response can then be altered to suit the needs of an expert designer, who may be interested in conducting specific design task and requires less subject-matter exploration. The objective is to eventually create an AI-assisted expert system to guide the user through the conceptual design phase. Some of the key advantages of such a system include:
\begin{itemize}
    \item Providing a methodological approach for conceptual design.
    \item Ability to search and fetch context-based information enabling intelligent workflow creation.
    \item Tailored to meet specific use cases while having the ability to cater to users of varying technical backgrounds.
\end{itemize}

\subsection{MyCrunchGPT for PINN-based simulations}
In addition to operator-based design, MyCrunchGPT can assist users in developing customized PINN models and codes for problems in scientific machine learning. The user describes the problem they want to solve using natural language, including relevant details about the PDE, domain, and any other relevant details or files needed to make the problem well-posed (e.g., observed data points, custom geometry files). MyCrunchGPT will formalize the problem the user describes, write and execute code, and provide various postprocessing of the solution if requested. The user can request to visualize different components of the solution, or verify the solution with existing solutions or other numerical codes. The user can request to modify the network architecture and other hyperparameters (optimizer, learning rate, weighting of boundary loss and interior loss, etc.), and MyCrunchGPT will modify and execute the code appropriately. A workflow diagram of the process is given in Figure \ref{fig:pinn_workflow}.

\subsection{MyCrunchGPT web application interface}
\begin{figure}[h]
  \centering
  \centerline{\includegraphics[width = 0.9\textwidth]{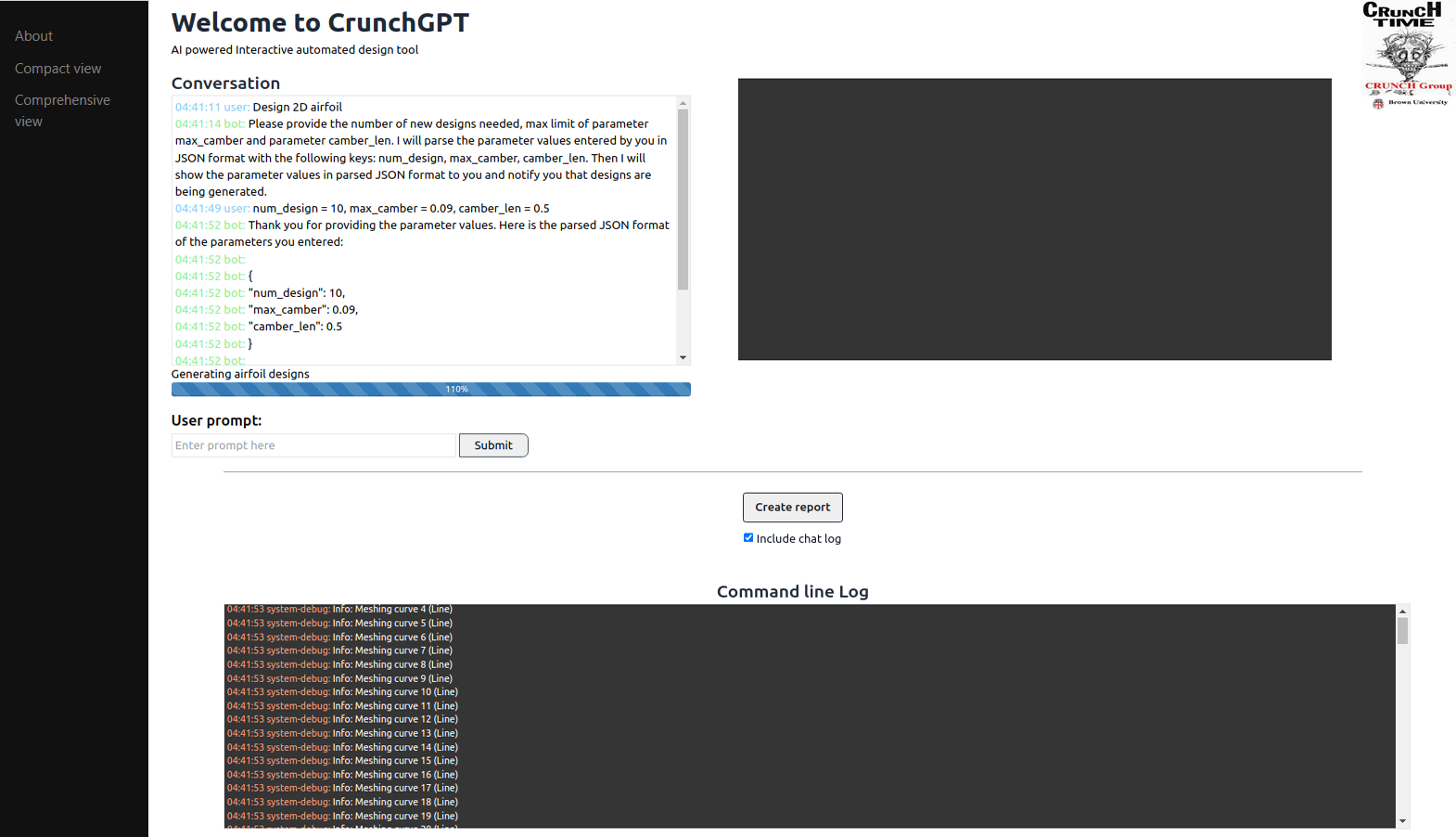}}
  \caption{Comprehensive view layout of MyCrunchGPT webapp. To the left, the navigation sidebar appears. In the main page the two panels are the conversation with the bot, and the images placeholder (in black). The user prompt box and the submit button appear under the conversation box. On the bottom we see a logger output with comprehensive logs. Last, between the upper and lower panels is the button for exporting the information on the screen into a PDF report, allowing the user to choose whether to include the chat or not.}
  \label{fig:CGPT_compview}
\end{figure}

To demonstrate the flow of the MyCrunchGPT, and create an  infrastructure that can facilitate the broader vision of this project, we built a web app based guided user interface (GUI). The GUI has a side bar for navigation that includes three pages: a) ``About'', transferring the user to a short description of the MyCrunchGPT, b) ``Compact view'', where the non-technical user can interact with the MyCrunchGPT, and c) ``Comprehensive view'', which is similar to the Compact view, but includes a console output view that projects in-depth debugging logs (serving ones that are interested in back-end logs). The key components of the GUI are:
\begin{itemize}
    \item \textbf{Conversation}: A view that projects the conversation with the bot. It shows the user prompts, the bot answers, and important system information (such as beginning and ending of long tasks). To achieve that we added to all the back-end codes custom log handling, with level management based on importance. Then, the GUI translates the messages and cherry-picks the ones to show in the conversation box.
    \item \textbf{Input box}: A text box with a submit button, which the user uses to interact with the model. This button is implemented using asynchronous threading, so that important information will appear on the screen while it is computed in the back-end. Webapps usually render information after a function call, while most back-end processes of the MyCrunchGPT take long time to compute. We had to apply the asynchronous method for information to appear mid-computation.
    \item \textbf{Image gallery}: An image viewer that can hold multiple images, and the user can scroll to view the images. This enables an interactive session, where the user can choose which images will be presented (and later written into the report), by visually examining the images.
    \item \textbf{Console} (only in Comprehensive view): A box showing back-end logs, such as iterations, system information, warnings and errors. These have been cherry-picked using the custom logs handler, to surface the functionality from the back-end, but not overflow the GUI with unnecessary logs.
    \item \textbf{Create report}: After the interaction is complete, we implemented a feature that allows the user to export the interactive session into a report of Portable Document Format (PDF) format. The user can choose whether to include the chat logs or not. The images and final recommendation, as well as justification, are included in the report by default.
\end{itemize} 
Figure \ref{fig:CGPT_compview} shows an overview of the different elements of the MyCrunchGPT webapp interface. 

\section{Exemplar for MyCrunchGPT assisted Operator-based Design framework} 
We demonstrate the application of MyCrunchGPT as a design assistant using the 2D NACA airfoil design and optimization problem \cite{shukla2023deep}. The problem statement is described as follows:

\begin{tcolorbox}[width=\textwidth, colback={white},title={\textbf{Problem statement}}, colbacktitle=lightgray, outer arc=2mm ,colupper=black, coltitle=black]  
   Create 2D NACA 4 digit airfoil designs based on maximum camber ($m$) and position of maximum camber ($p$) specified by the user. Generate several new designs as requested by the user. Once generated, analyze the flow field around these new designs. Also, generate an optimized airfoil geometry within the range of parameters $m$ and $p$ defined by the user.
\end{tcolorbox}   

This problem was chosen as an example to demonstrate the MyCrunchGPT integrated design process workflow due to availability of existing DeepONet surrogate and CFD simulation data \cite{shukla2023deep}. We acknowledge the need to extend our MyCrunchGPT framework to other design problems and scenarios and consider this as a future work. Here, we describe the design problem setup briefly, by dividing the overall design problem into smaller subsets for clarity.

\subsection{Problem subset 1: Generating new airfoil designs} \label{section:gen_design}
The surface of the airfoils are generated using the parametric equations with a chord length of one, given as:
\begin{align} \label{eqn: airfoil_param}
& y_t=\frac{t}{0.2}\left(a_0 \sqrt{x}+a_1 x+a_2 x^2+a_3 x^3+a_x x^4\right) \\
& y_c= \begin{cases}\frac{m}{p^2}\left(2 p x-x^2\right) & \text { if } x<p \\
\frac{m}{(1-p)^2}\left(1-2 p+2 p x-x^2\right) & \text { if } x>p\end{cases} \\
& \theta=\tan ^{-1}\left(\frac{d y_c}{d x}\right) \\
& x_u=x_c-y_t \sin (\theta), \quad y_u=y_c+y_t \cos (\theta) \\
& x_l=x_c+y_t \sin (\theta), \quad y_l=y_c-y_t \cos (\theta)
\end{align}
where $ a_{0} = 0.2969 $, $ a_{1} = -0.1260 $, $ a_{2} = -0.3516 $, $ a_{3} = 0.2843 $, $ a_{4} = -0.1015 $. The 2D airfoil geometry is therefore defined as a point cloud with coordinate sets $(x_u, y_u, x_l, y_l)$ parameterized by $\eta_g = (t,p,m)$. As a further simplification to reduce likelihood of flow separation and turbulence effects, the parameter space $\eta_g$ is constrained as follows: the maximum thickness ($t$) is set to a constant value of 0.15, and the domain of the parametric space left by the position of maximum camber ($p$) and maximum camber ($m$) is $p \times m \in [0.20, 0.50] \times [0.01, 0.09]$. Here, 100 points are sampled along the x-direction, leading to a set of 100 points \{$x_{coord},y_{coord}$\} that define the coordinates of the airfoil surface. To reduce the dimensionality of the input space for training DeepONet surrogate, Non-Uniform Rational
B-Splines (NURBS) with 30 control points are used, hence reducing the input data dimension to 30. 

\subsection{Problem subset 2: Analyzing flow fields around new designs}
\begin{figure}[h]
  \centering
  \centerline{\includegraphics[width = 0.8\textwidth]{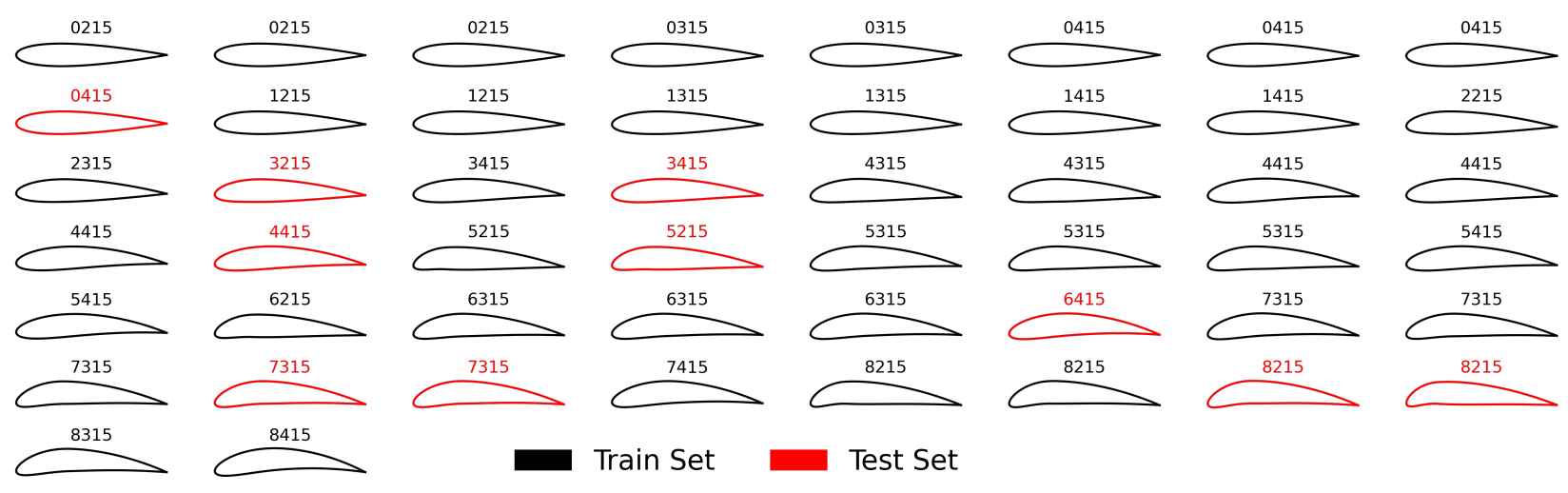}}
  \caption{DeepONet surrogate training and testing geometries generated from random sampling from parameter space $\eta_g$. Note that the numbers correspond to NACA airfoils and duplicate numbers are due to rounding to the nearest integer for readability. The true NACA parameters are drawn from a uniform distribution and are real-valued. Adapted from \cite{shukla2023deep}.}
  \label{fig:deeponet_training}
\end{figure}

CFD simulations are widely used for analyzing fluid flow within a domain of interest. Compressible fluid flow simulations using viscous Navier-Stokes equations are often used to study the effect of fluid medium on an object moving through it at supersonic speeds. However, traditional CFD methods using compressible flow solvers are computationally expensive and analyzing several designs simultaneously is a major bottleneck in design exploration phase. As a workaround, fast and accurate surrogate models that can work in tandem with expensive solvers provide a promising alternative, especially in the conceptual design phase where multiple design concepts need to be analyzed. In previous work \cite{shukla2023deep}, an operator-based neural network (DeepONet) surrogate model was developed and tested as an alternative to CFD simulation on a set of 2D NACA airfoil designs. The DeepONet surrogate is trained using CFD simulation data generated for forty 2D NACA airfoil designs using Nektar++ \cite{cantwell2015nektar}. We leverage this pre-trained DeepONet surrogate for analyzing new designs generated through the MyCrunchGPT design assistant. Details for the CFD setup can be found in \cite{shukla2023deep}, and figure \ref{fig:deeponet_training} shows the training and testing geometries used for the DeepOnet surrogate training.

DeepONet surrogate enables one to map from functional space to operator space, which requires domain sub-sampling. In this work, this sub-sampling is accomplished by generating a mesh inside the fluid domain using the free mesh generator tool, Gmsh \cite{geuzaine2009gmsh}. We use mesh refinement to reduce the mesh size around the leading and trailing edge of the airfoils. The mesh generation is automated using Gmsh's Python API, which is integrated with the design parametrization step discussed in section \ref{section:gen_design}. 

\subsection{Problem subset3: Generate optimized airfoil geometry}
Shape optimization for geometric shapes is an essential step in design process, where an objective function is minimized/maximized based on a set of constraints. Here, the optimization problem is to maximize the lift-to-drag ratio over a feasible region of parameters, which are $m$ and $p$ in our problem. The optimization problem is stated as:
\begin{align}
\begin{array}{cl}
\underset{m, p}{\operatorname{minimize}} & -LD(m, p) \\
\text { subject to } & m_{\min } \leq m \leq m_{\max } \\
& p_{\min } \leq p \leq p_{\max }
\end{array}
\end{align}
where $LD(m,p)$ represents the lift to drag ratio. This lift ($L$) and drag ($D$) are calculated using equation \ref{eqn: shear_str} - \ref{eqn: Drag}. The map created by DeepONet surrogate takes an input geometry and spatial (x, y) location in the output space and returns the value of the field at that location. This allows estimation of lift and drag through discrete integral of flow fields over the surface of the airfoil
\begin{align}
\tau_w & =\mu (\nabla \bm{u} + \nabla \bm{u}^\top) \cdot \bm{n}\label{eqn: shear_str}\\ 
D & =\int d F_x=\sum p \hat {n}_{x} d S+\sum \tau_w \hat{t}_{x} d S \label{eqn: lift} \\ 
L & =\int d F_y=\sum p \hat {n}_{y} d S+\sum \tau_w \hat{t}_{y} d S .  \label{eqn: Drag}  
\end{align}

For calculating gradients of output fields with respect to coordinate locations inside the domain, we use automatic differentiation of computational graph creating during DeepONet evaluation. The analytical derivative $\dfrac{d U}{d \hat{n}}$ is given by:
\begin{equation}
\frac{d U}{d \hat{n}}=\left(v_y-u_x\right) \sin \theta \cos \theta-v_x \sin ^2 \theta+u_y \cos ^2 \theta
\end{equation}
where the partial derivatives of the velocity field $\{u_x, u_y, v_x, v_y\}$ are evaluated using automatic differentiation.

For optimization, we use a simulated dual-annealing method inside python library SciPy. Dual-annealing iteratively searches for a potential solution based on an acceptance criteria and draws inspiration from simulated annealing optimization technique \cite{simannealing}. The process for generating an optimized airfoil shape is as follows:
\begin{enumerate}
    \item [Step 1:] Random sampling from parameter space $p \times m \in [0.20, 0.50] \times [0.01, 0.09]$ using Latin hypercube. These parameters are used as inputs to branch network.
    \item [Step 2:] Airfoil surface profile generation using equation \ref{eqn: airfoil_param}. Thereafter, NURBS curve fitting to airfoil geometry to define the second input to branch network.
    \item [Step 3:] Generation of flow field using the DeepONet surrogate using trunk's input coordinates. Calculation of lift and drag from the flow field values.
    \item [Step 4:] Optimization step, with new parameter selection in next iteration. Repeat until convergence or user defined step is reached.
\end{enumerate}

\subsection{Use-case analysis}
\begin{figure}[h]
  \centering
  \centerline{\includegraphics[width = 0.8\textwidth]{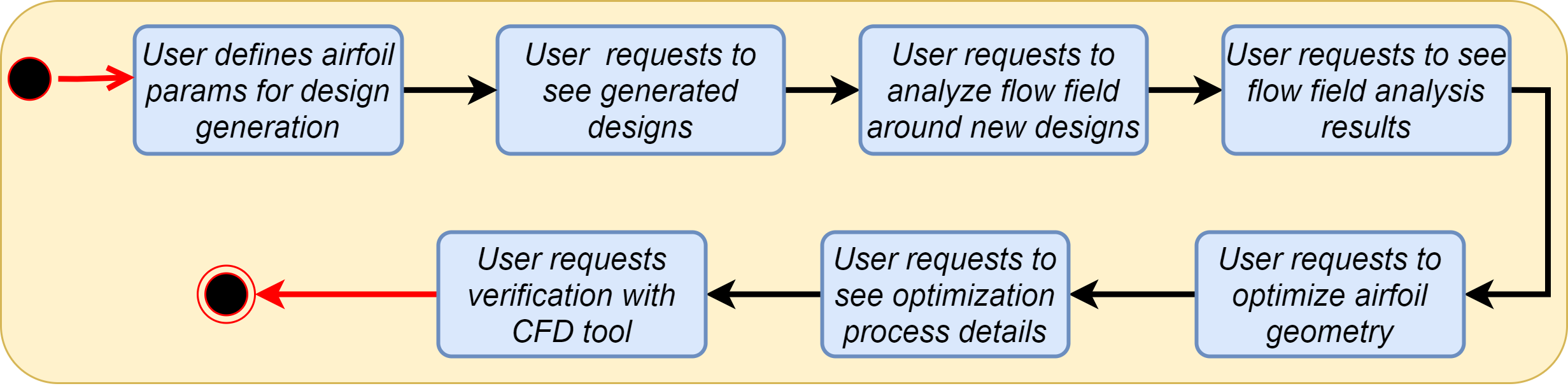}}
  \caption{Use case scenario for 2D airfoil design. The scenario contains expected steps that a user will undertake during a 2D airfoil design process. MyCrunchGPT responds as an assistant during this process to guide the designer while collecting necessary information required for completing user requests. The MyCrunchGPT assistant is instructed to respond to a user entry with an appropriate human-like response.}
  \label{fig:use_case_deeponet}
\end{figure}

The use case scenario considered in our MyCrunchGPT Design assistant example is shown in figure \ref{fig:use_case_deeponet}. The user requirements for this design process include:
\begin{itemize}
    \item Generating a chosen number of new airfoil designs using specified values of max camber ($m$) and its position ($p$). 
    \begin{tcolorbox}[width=0.8\textwidth, colback={white},title={\textbf{MyCrunchGPT role}}, colbacktitle=pink, outer arc=2mm ,colupper=black, coltitle=black]  
    collects user inputs and calls the appropriate python API to start design generation
\end{tcolorbox}  
    \item Analyzing the new airfoil designs using the DeepONet surrogate. 
    \begin{tcolorbox}[width=0.8\textwidth, colback={white},title={\textbf{MyCrunchGPT role}}, colbacktitle=pink, outer arc=2mm ,colupper=black, coltitle=black]  
    initializes the pre-trained DeepONet model, calls necessary APIs to evaluate the new designs through the DeepONet surrogate, and generates the flow fields
    \end{tcolorbox} 
    \item Viewing flow field results.
    \begin{tcolorbox}[width=0.8\textwidth, colback={white},title={\textbf{MyCrunchGPT role}}, colbacktitle=pink, outer arc=2mm ,colupper=black, coltitle=black]  
    parses user input appropriately, calls the visualization APIs to plot requested flow field
    \end{tcolorbox} 
    \item Generating an optimized airfoil geometry to minimize lift to drag ratio.
    \begin{tcolorbox}[width=0.8\textwidth, colback={white},title={\textbf{MyCrunchGPT role}}, colbacktitle=pink, outer arc=2mm ,colupper=black, coltitle=black]  
    calls the optimization module (using SciPY optimizer). The optimizer module samples the parameter space, creates new airfoil designs based on the sampled parameters, and optimizes the design based on objective function
    \end{tcolorbox} 
    \item Verifying DeepONet results against CFD tool.  
    \begin{tcolorbox}[width=0.8\textwidth, colback={white},title={\textbf{MyCrunchGPT role}}, colbacktitle=pink, outer arc=2mm ,colupper=black, coltitle=black]  
    initiates the CFD tool API (Nektar++), requests user for simulation settings, updates CFD simulation script, and initiates the CFD run.
    \end{tcolorbox} 
\end{itemize}

For demonstration, we simplify the use case with a step-by-step process flow diagram (Figure \ref{fig:use_case_deeponet}), although we note that the GPT-based design assistant can learn tasks that are more complex, even cyclical due to the ability of chatGPT model to understand and learn long command sequences. 

\subsection{Workflow for DeepONet-based airfoil design}
\begin{figure}[h]
  \centering
  \centerline{\includegraphics[width = 0.5\textwidth]{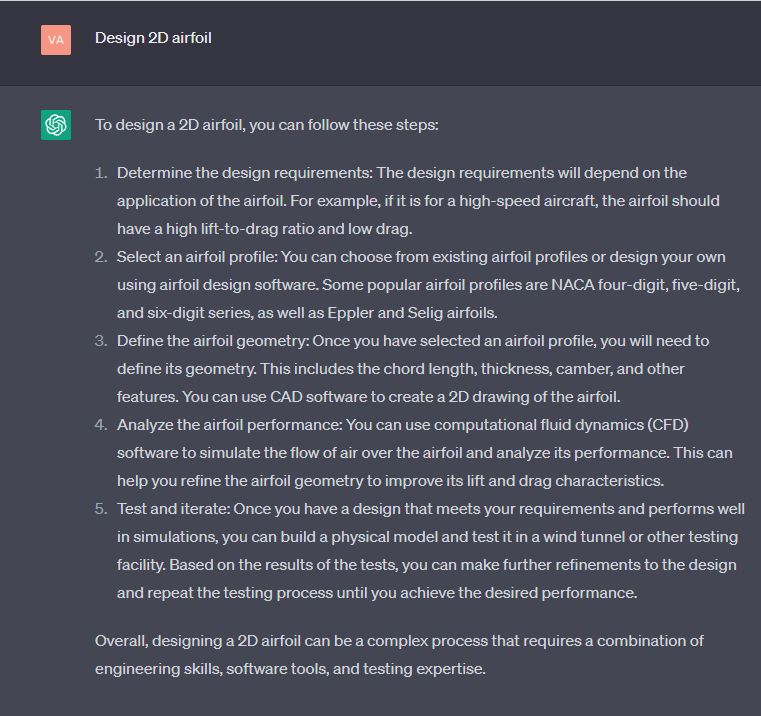}}
  \caption{Generic responses generated by chatGPT when the GPT assistant is not primed with instructions for a specific use case. Here, when a user asks chatGPT to help with 2D airfoil design, a generic response is received. However, for a specific task, such generic responses may not lead to actionable feedback for the designer. Note that this response is generated by the chatGPT web-based interface (\url{https://chat.openai.com}).}
  \label{fig:gpt_generic_response}
\end{figure}

Based on the use case scenario for our MyCrunchGPT, we establish the process workflow required for the design assistant. A necessary step for customizing chatGPT for the purpose of acting as a design assistant involves creating a sequence of instructions that MyCrunchGPT needs to adhere to when responding to user requests. This is critical since chatGPT (which forms the backbone for our MyCrunchGPT assistant) is a large language model, and produces generic information. For instance, see figure \ref{fig:gpt_generic_response} where chatGPT responds to a user query seeking help with 2D airfoil design with a set of generic guidelines and steps that may be taken. However, this information may not be useful for effective action from a designer. 

To receive task-specific responses from chatGPT, we create a set of instructions to modulate the response based on our use case of the 2D airfoil design. These instructions are appended to the message history using chatGPT 3.5 API, and set the manner in which chatGPT should respond to user queries. As an example, we add the following specific instruction (see Figure \ref{fig:instructions_eg}) to chatGPT API's message history asking the GPT server to request user for specific parameters (max\_camber, $m$ and position of max\_camber, $p$) when the user requests for generation of 2D airfoil design:

\begin{figure}[h]
  \centering
  \begin{tcolorbox}[width=0.9\textwidth, colback={white},title={\textbf{MyCrunchGPT instructions}}, colbacktitle=yellow, outer arc=2mm ,colupper=black, coltitle=black]  
   ["role": "user", "content": ``You are an assistant to a design engineer. When user requests to design 2D airfoil, prompt user to provide number of new designs needed, max limit of parameters max camber and camber len. Only ask for these values in your response. Then notify user when designs are being generated.'']
\end{tcolorbox}
  \caption{Example of instruction used to tailor chatGPT's response to specific task of designing 2D airfoil based on user query. These instructions are fed to the message history inside chatGPT thus providing the language model with context for these instructions. This enables us to tune the chatGPT response to be specific to our tasks.}
  \label{fig:instructions_eg}
\end{figure}

\begin{figure}[h]
  \centering
  \centerline{\includegraphics[width = 0.9\textwidth]{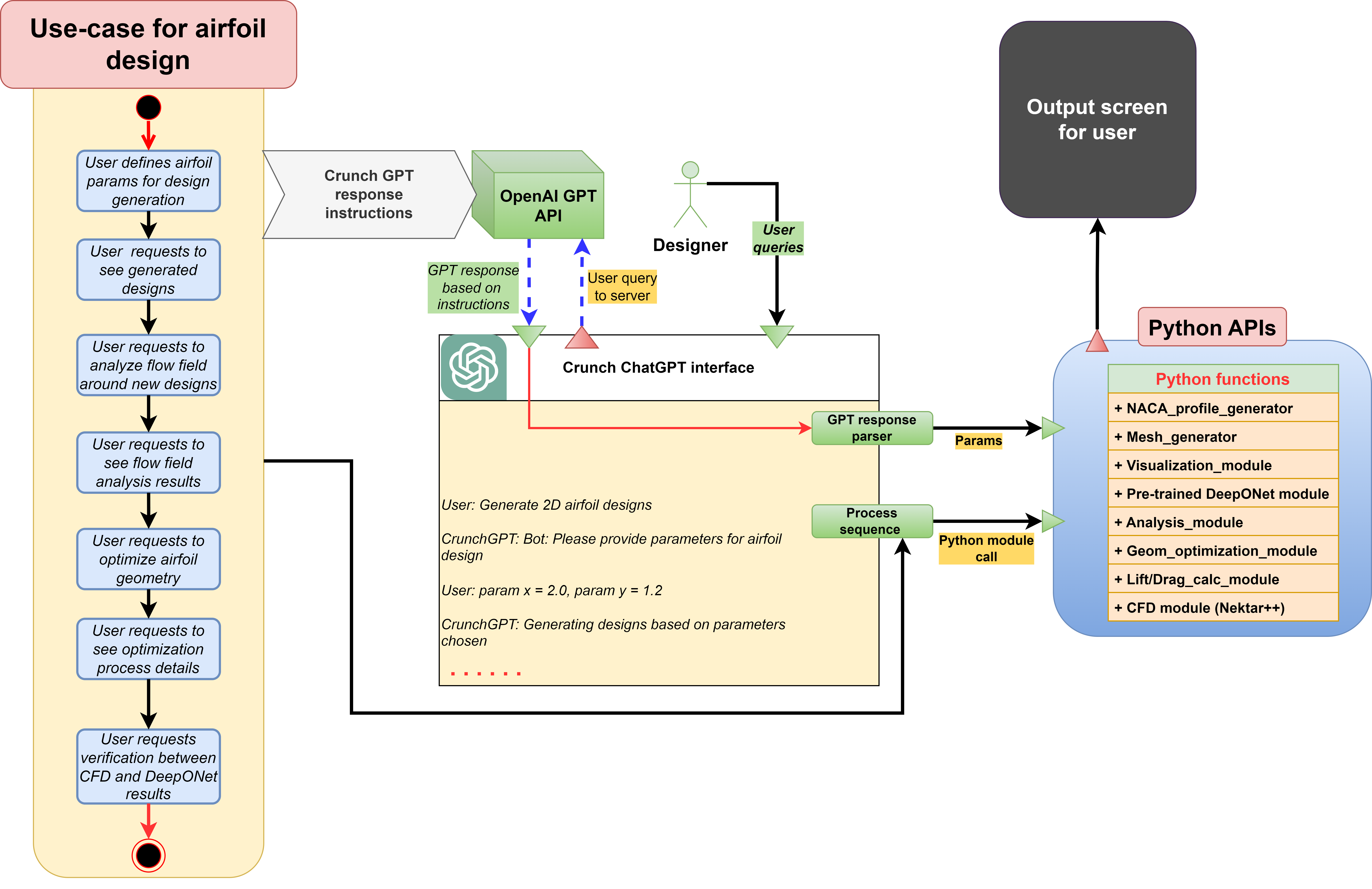}}
  \caption{Schematic showing workflow for MyCrunchGPT design assistant. Instructions for priming MyCrunchGPT are determined by the sequence of steps outlined in the use-case for airfoil design, as shown on the left side here. These instructions enable us to receive specific response to a user query. The user responses are parsed into required parameters for performing the actions necessary by calling respective python modules.}
  \label{fig:workflow_gpt_design}
\end{figure}

Since chatGPT is a sequential learning language model, a sequence of such instructions can be passed to chatGPT API, thereby enabling specific responses to meet the use case requirements. This sequential learning also helps to maintain context for GPT, and this ability is useful for managing complex workflows. Since chatGPT API currently does not have the ability to perform operations such as read/write or open files on a user's computer, we use pre-defined python functions to accomplish the necessary tasks. The sequence in which these functions are called is determined by the use case scenario shown in figure \ref{fig:use_case_deeponet}. The complete workflow for Crunch GPT for design is shown in figure \ref{fig:workflow_gpt_design}. Using the workflow as described in figure \ref{fig:use_case_deeponet}, a set of instructions are generated to prime MyCrunchGPT to respond to user queries. A parsing operation parses responses from chatGPT to extract relevant parameters to execute the next course of action. For instance, when the user enters the parameters $m$ and $p$, the chatGPT parses this information into a JSON format, which is processed by the parser operation to generate parameters necessary to call the necessary python operation (`NACA profile generator') in this case. MyCrunchGPT performs the sequence of steps as requested by the user, seeking necessary information and providing appropriate feedback to the user at each stage of the design process. 

We demonstrate the response generated by the MyCrunchGPT design assistant at each step in the following section by using screenshots from the user interface.

\begin{figure}[h]
  \centering
  \centerline{\includegraphics[width = 0.7\textwidth]{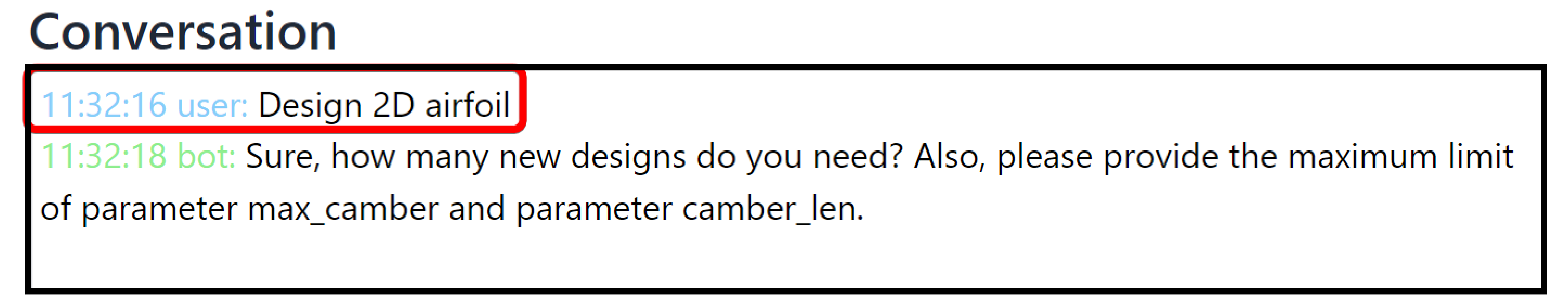}}
  \caption{\textbf{[Request 1]} MyCrunchGPT's response to user query to design 2D airfoil asking for parameter values $m$ and $p$ as well as total number of new designs that are required. Compare how the response has been tailored for our use-case vs a generic response generated by chatGPT shown in figure \ref{fig:gpt_generic_response}.}
  \label{fig:prompt1}
\end{figure}

\textbf{Request1 -> User request to design 2D airfoil:} MyCrunchGPT requests the user to provide values for relevant design paramaters based on its pre-defined instruction (Figure \ref{fig:prompt1}).

\begin{figure}[h]
  \centering
  \centerline{\includegraphics[width = 0.65\textwidth]{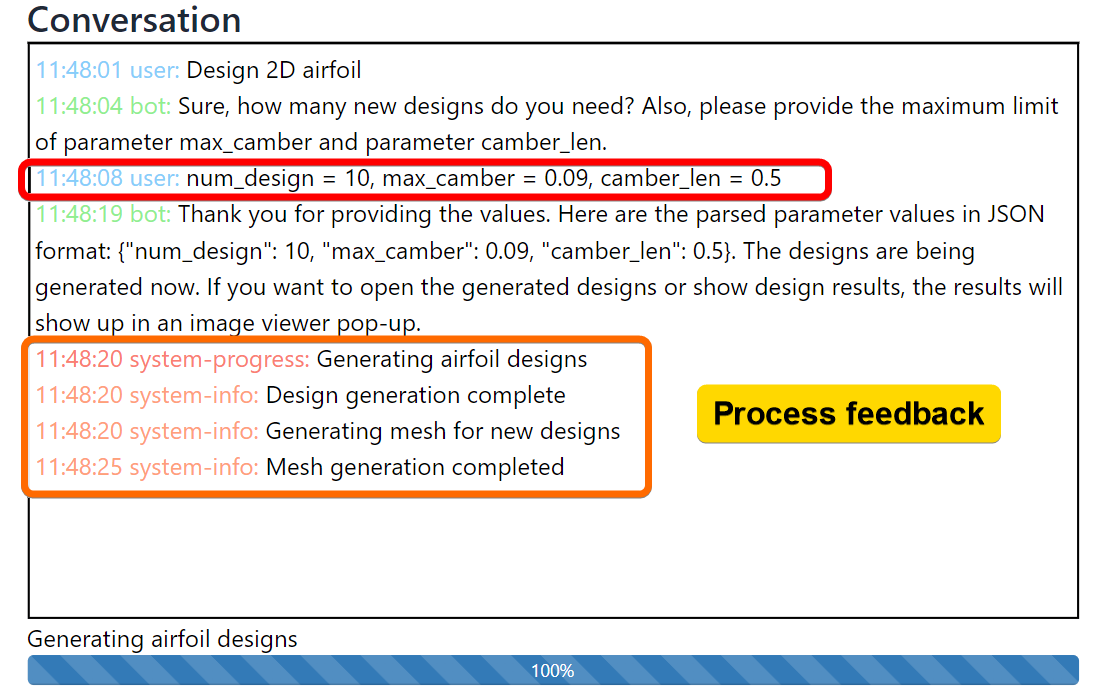}}
  \caption{\textbf{[Request 2]} Parameter values provided by user which are parsed by MyCrunchGPT to provide feedback to user and initiate new airfoil design process by invoking the necessary python function. MyCrunchGPT provides continuous process feedback to the user about progress.}
  \label{fig:prompt2}
\end{figure}

\textbf{Request2 -> User provides requested parameters:} MyCrunchGPT notifies user about the parameter values provided and that the design process is initiating. In the background, the required python function (`NACA mesh generator') is used for generating new design by parsing the parameter values provided by the user; see Figure \ref{fig:prompt2}.

\begin{figure}[h]
         \centering
         \centerline{\includegraphics[width = 1\textwidth]{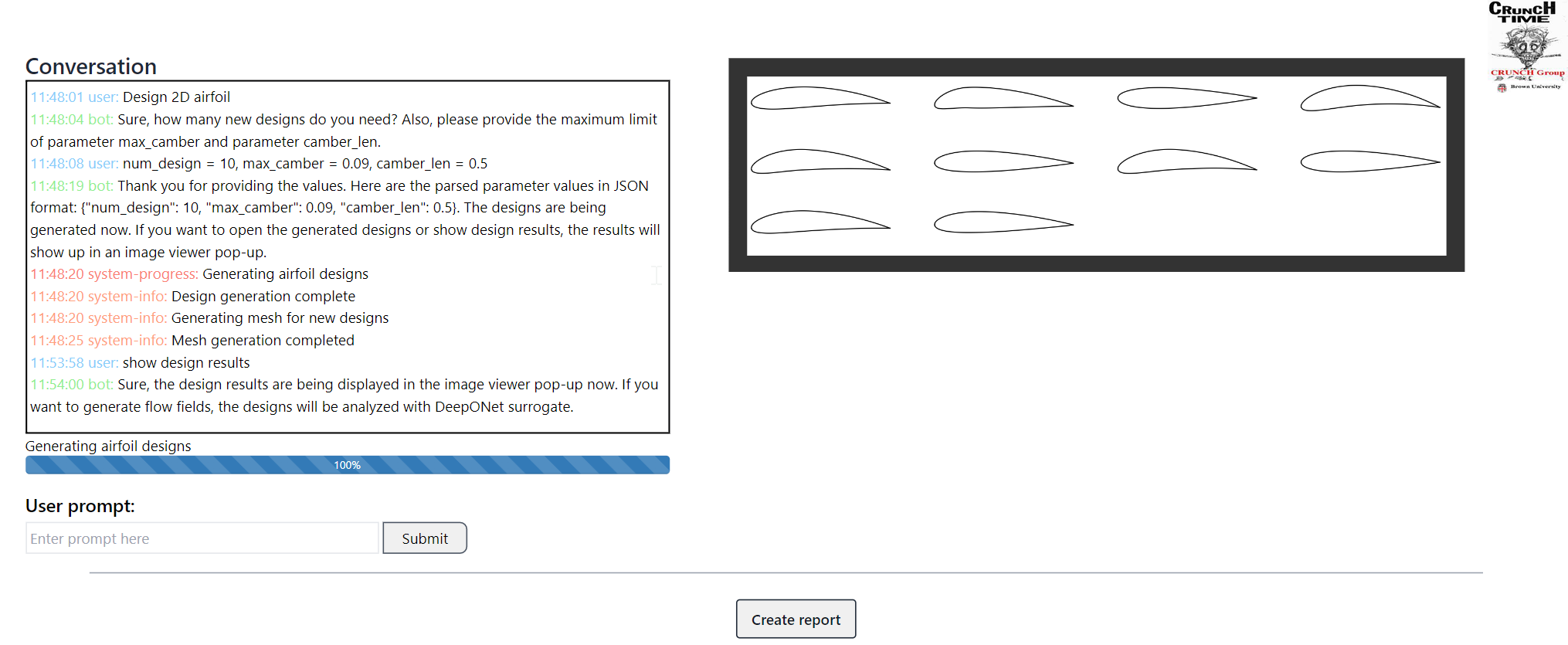}}
       \caption{\textbf{[Request 3]} User requests MyCrunchGPT to show new airfoil designs generated in the design process. Newly generated airfoil designs  are displayed in the webapp's image viewer.}
       \label{fig:design_results}
\end{figure}

\textbf{Request3 -> Display results:} When the user requests MyCrunchGPT to display generated designs, the visualization routine is called in the back-end to display the new airfoil designs generated. The results appear on the webapp's image viewer window; see Figure \ref{fig:design_results}.

\begin{figure}[h]
         \centering
         \centerline{\includegraphics[width = 0.6\textwidth]{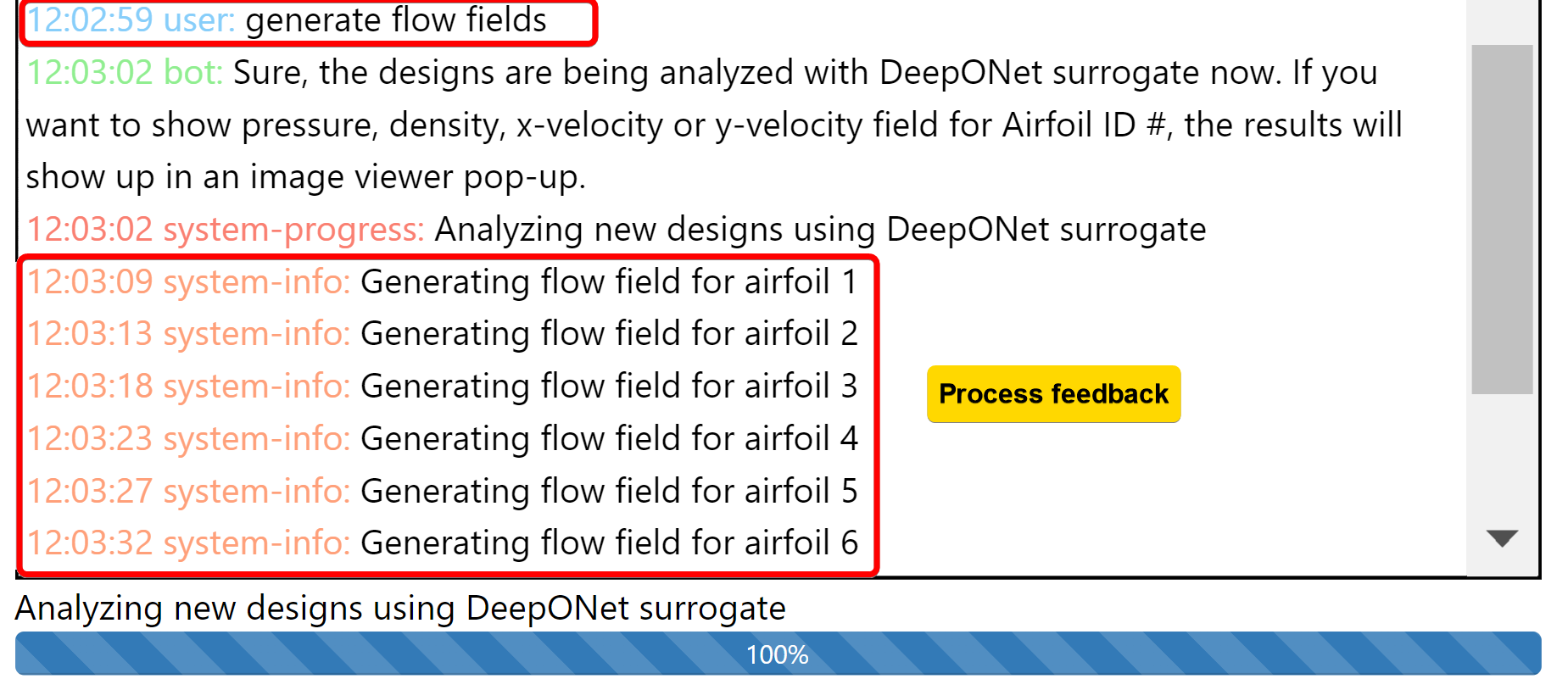}}     
       \caption{\textbf{[Request 4]}  User requests MyCrunchGPT to analyze flow field around newly designed airfoils. Using this request, MyCrunchGPT calls the necessary python function for analyzing flow fields by calling pre-trained DeepONet. This process takes some time depending on the number of new designs to be evaluated but is significantly less than running a CFD solver. User is provided with continuous feedback while the new designs are being evaluated through the pre-trained DeepONet.}
       \label{fig:don_simulation}
\end{figure}

\textbf{Request4 -> Generate flow fields:} When the user requests MyCrunchGPT to analyze flow fields around the new airfoil designs, the designs are analyzed using the surrogate DeepONet in the back-end (\cite{shukla2023deep}). This surrogate generates the flow field around the new airfoil designs by calling the necessary python function (Figure \ref{fig:don_simulation}). The user is notified once the design analysis process is completed.  

\begin{figure}[h]
         \centering
         \centerline{\includegraphics[width = 1\textwidth]{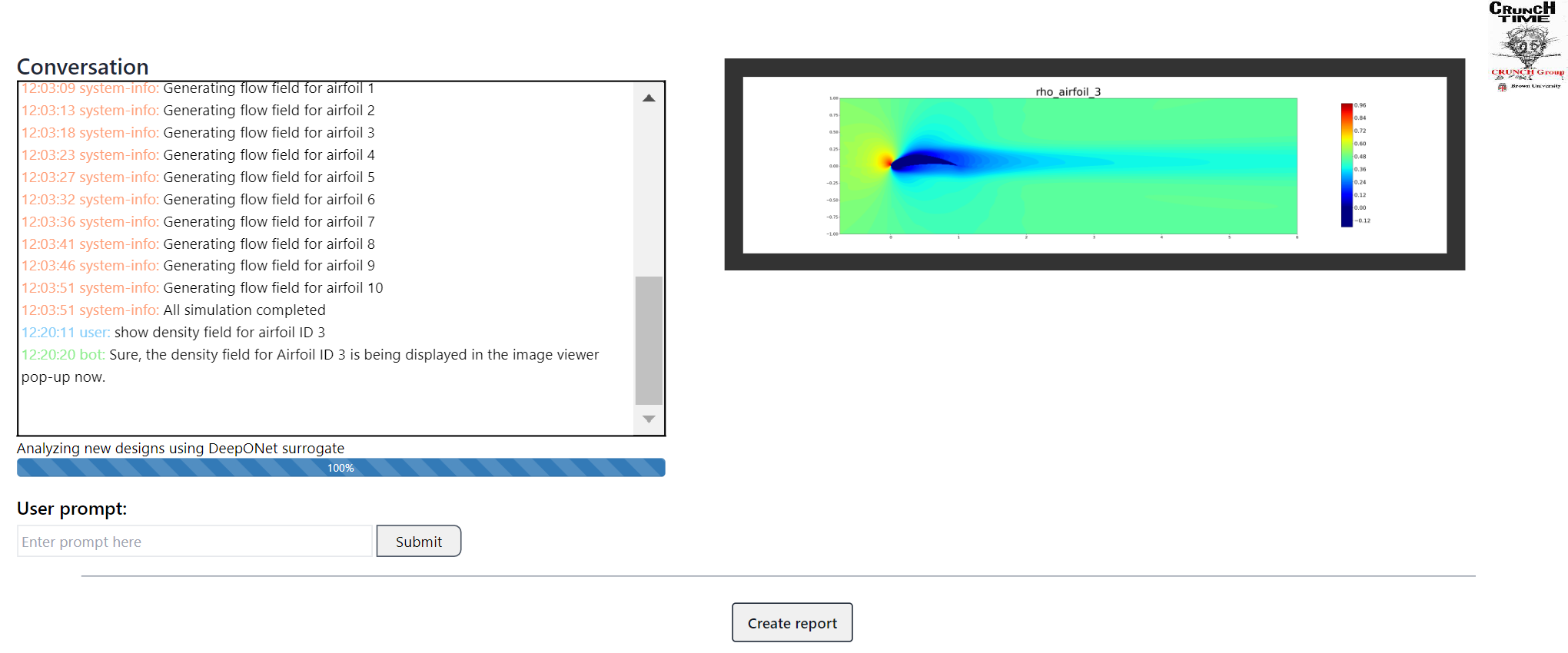}}
       \caption{\textbf{[Request 5]} User requests MyCrunchGPT to show density flow field around airfoil 3. The requested density flow field is plotted and displayed.}
       \label{fig:rho_field}
\end{figure}

\textbf{Request5 -> Show flow fields:} A user, after looking at the airfoil designs may wish to check the flow field around a particular design. Upon receiving such a request, MyCrunchGPT invokes a visualization routine in the back-end and the necessary flow field is displayed in the image viewer (figure \ref{fig:rho_field}). 

\begin{figure}[h!]
     \centering
     \begin{subfigure}[b]{0.47\textwidth}
         \centering
         \centerline{\includegraphics[width = 1\textwidth]{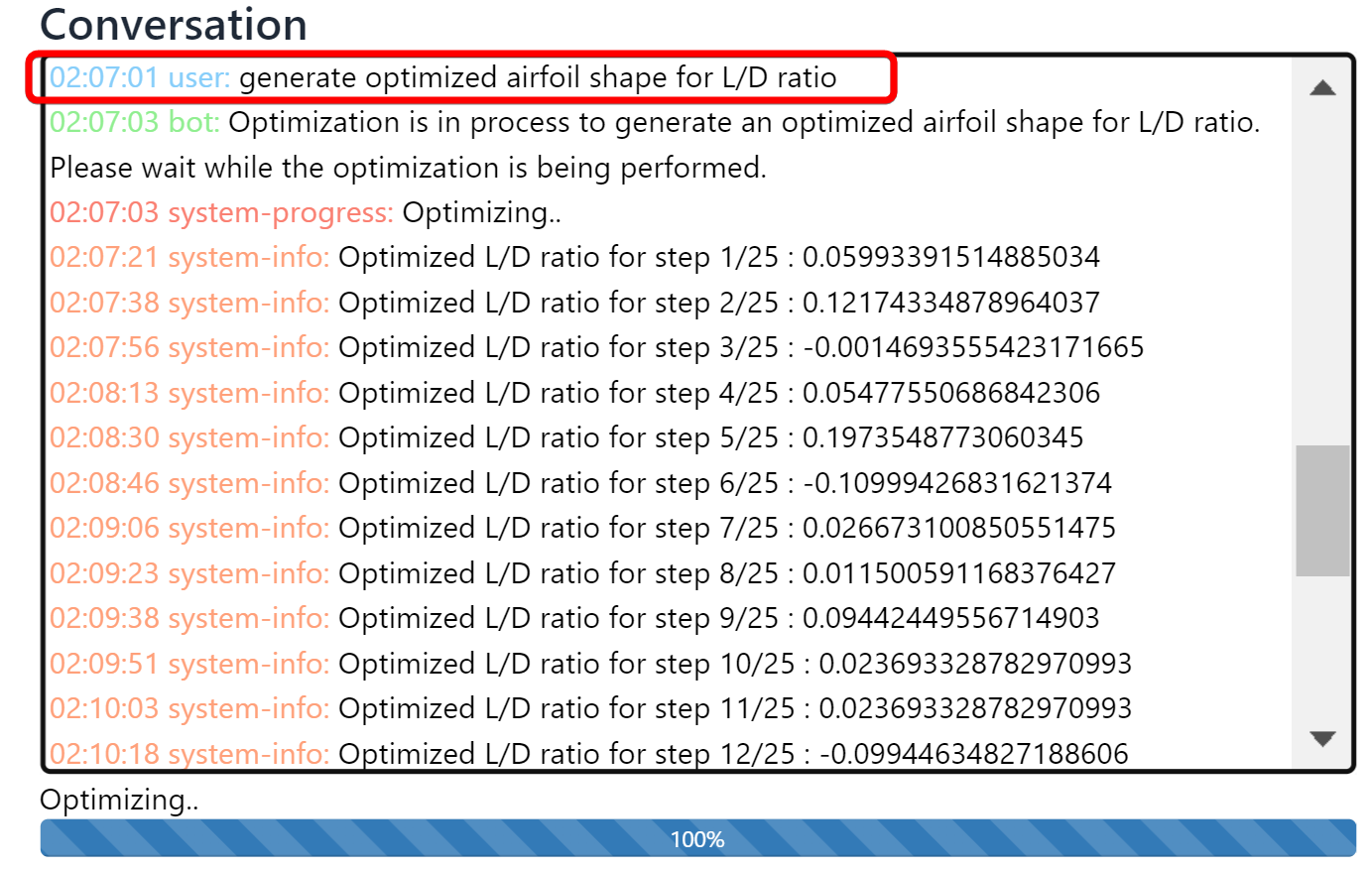}}
         \caption{}         
     \end{subfigure}
     \hfill
      \begin{subfigure}[b]{0.5\textwidth}
         \centering
         \centerline{\includegraphics[width = 1\textwidth]{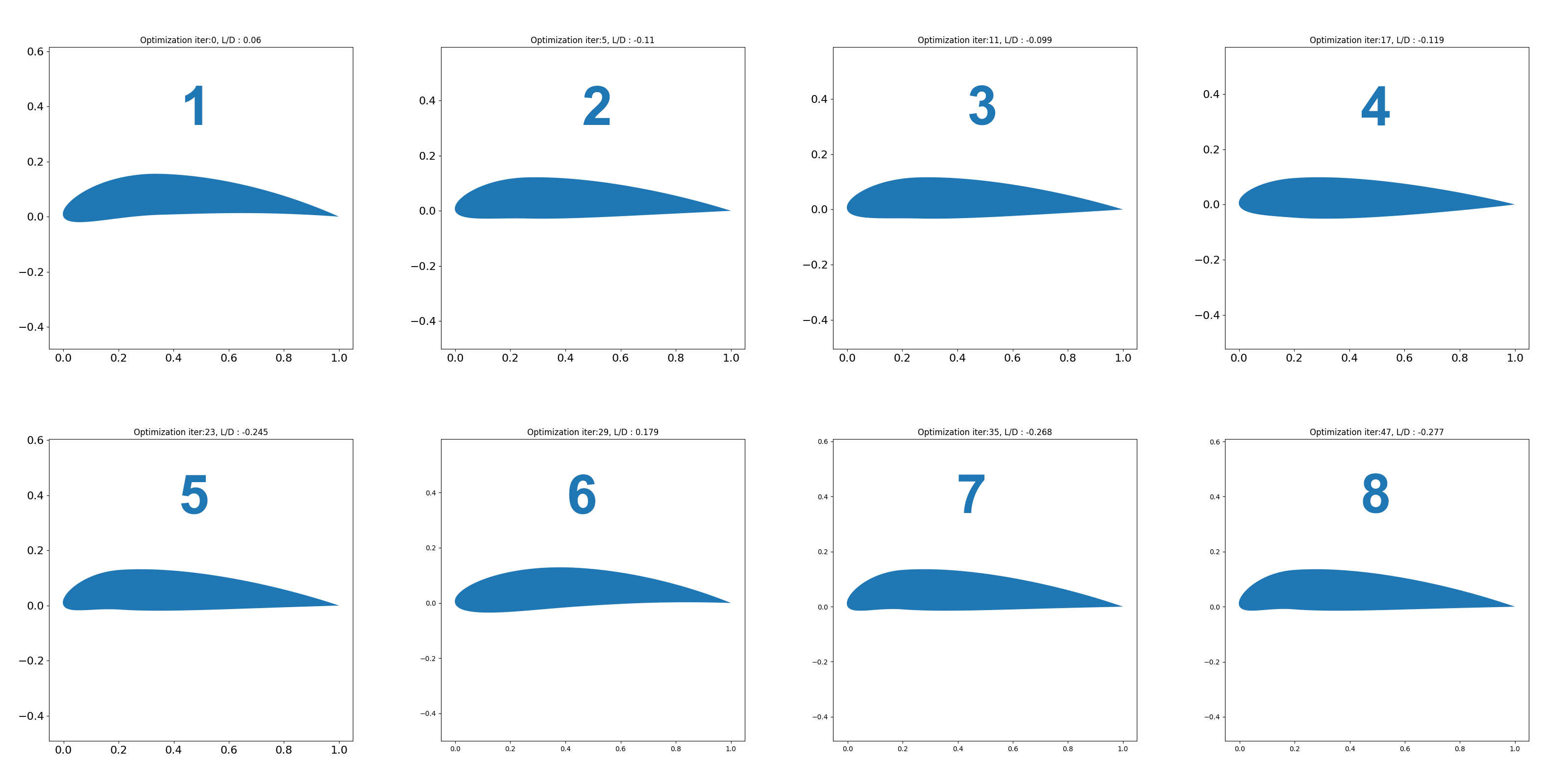}}
         \caption{}         
     \end{subfigure}
       \caption{(a) \textbf{[Request 6]}  User requests MyCrunchGPT to initiate optimization process to generate the best profile to maximize lift to drag ratio. MyCrunchGPT calls necessary python modules in the back-end and initiates the optimization process, providing regular process updates through an interface showing how the design is evolving with each iteration. (b) Airfoil geometries being updated during the optimization process displayed to user.}
       \label{fig:optimization}
\end{figure}

\textbf{Request6 -> Generate optimized airfoil shape:} When user requests MyCrunchGPT to create an optimized airfoil geometry to minimize lift-to-drag ratio, MyCrunchGPT calls the optimization module in back-end that initiates the process. User is notified about the progress through an interface window that shows the evolving airfoil shape with each iteration. The number of optimization iterations required can also be defined by the user (not shown in this example). User is notified of the optimization progress through MyCrunchGPT interface (Figure \ref{fig:optimization}).

\begin{figure}[h!]
     \centering
         \centerline{\includegraphics[width = 0.9\textwidth]{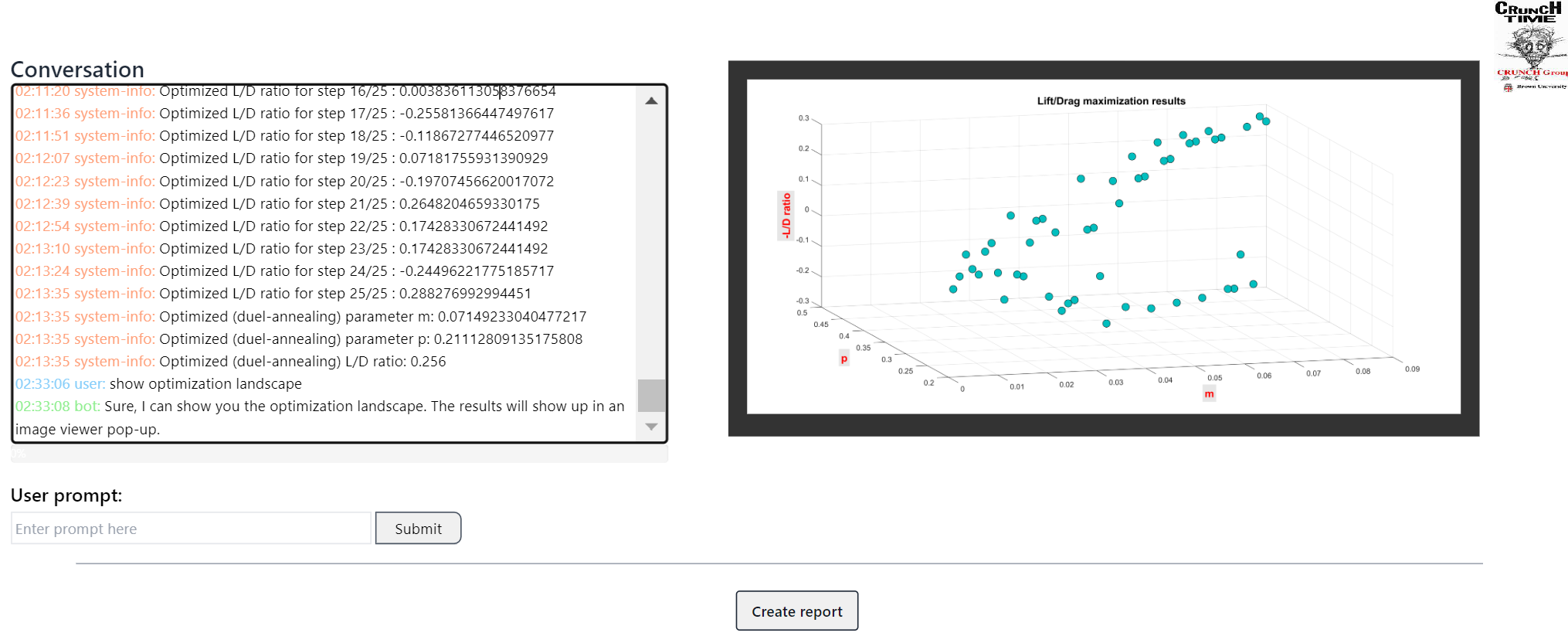}}
       \caption{\textbf{[Request 7]} User requests MyCrunchGPT to show the optimization landscape to analyze design space exploration during optimization. MyCrunchGPT invokes the visualization module to display the required plot to user in the image viewer.}
       \label{fig:optim_landscape}
\end{figure}

\textbf{Request7 -> Show optimization landscape:} When the user requests MyCrunchGPT to show the optimization landscape for analysis, MyCrunchGPT invokes the visualization module in the back-end to generate necessary plots showing how design is evolved during the optimization process; see Figure \ref{fig:optim_landscape}.

\begin{figure}[h!]
  \centering
  \centerline{\includegraphics[width = 0.6\textwidth]{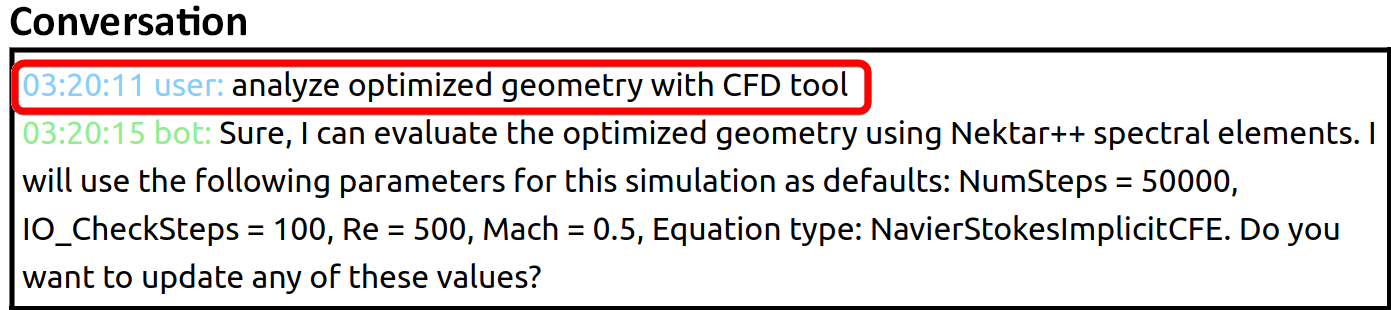}}
  \caption{\textbf{[Request 8]} User request MyCrunchGPT to analyze the optimized airfoil geometry with a CFD tool. MyCrunchGPT then provides the user with the default settings that will be used for the CFD solver, such as number of iteration steps (NumSteps). MyCrunchGPT then waits for the user's response to determine if the CFD solver settings need to be modified or not.}
  \label{fig:prompt7}
\end{figure}

\textbf{Request8 -> Verify DeepONet results for optimized airfoil against CFD tool:} As a verification step, a user may want to evaluate the optimized geometry with a CFD tool. MyCrunchGPT has an opensource CFD tool, Nektar++ integrated in its framework, which can be invoked upon user's request. At this prompt, MyCrunchGPT notifies the user about the CFD tool it can use, the default parameter settings that will be used for the CFD simulation (Figure \ref{fig:prompt7}). Additionally, MyCrunchGPT checks with the user to see if any of these settings need to be modified as per user's needs.

\begin{figure}[h!]
     \centering
 \begin{subfigure}[b]{0.45\textwidth}
         \centering
         \centerline{\includegraphics[width = 1\textwidth]{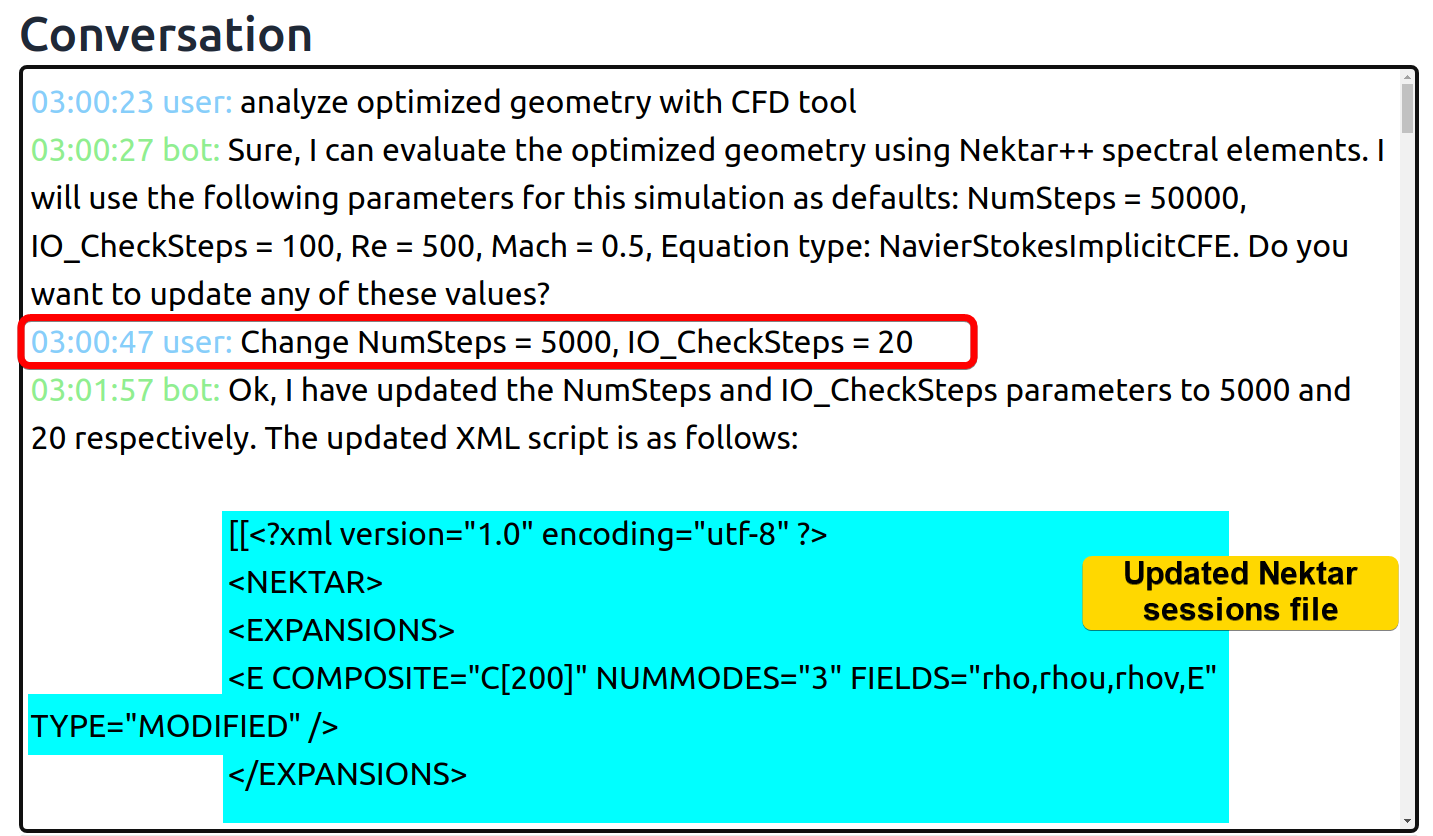}}
         \caption{} 
         \label{fig:prompt8}
     \end{subfigure}     
     \hfill
      \begin{subfigure}[b]{0.45\textwidth}
         \centering
         \centerline{\includegraphics[width = 1\textwidth]{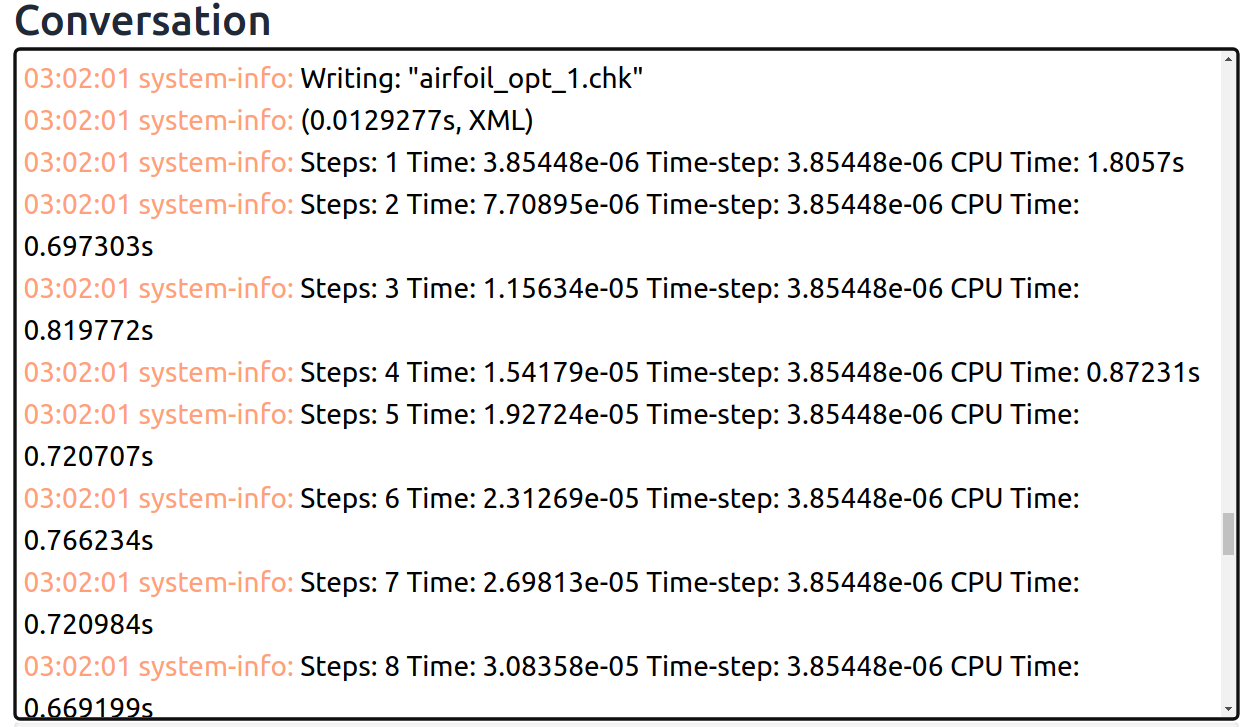}}
         \caption{}
         \label{fig:prompt8_out}
     \end{subfigure}
     \caption{(a) Updated CFD settings generated by MyCrunchGPT based on changes requested by the user. MyCrunchGPT reads the default CFD settings file provided through GPT instructions, identifies the correct fields (\texttt{`NumSteps'} and \texttt{`IO\_CheckSteps'} in this case) that need to be changed and then updates it with the corresponding user input. Thereafter, MyCrunchGPT displays the updated settings for CFD simulation to the user for user verification (complete MyCrunchGPT modified simulation settings script not shown here for brevity). (b) Once CFD settings are confirmed by user, MyCrunchGPT invokes the Nektar++ API to initiate simulation with constant feedback about simulation progress.}
\end{figure}

Once the user enters the values of the updated parameters for CFD simulation, MyCrunchGPT then parses this information, analyzes the default simulation settings file, identifies the parameters in this file that need to be changed, updates these parameters with user defined values, and then displays the updated code to the user for verification (Figure \ref{fig:prompt8}). Once the user confirms these CFD settings, MyCrunchGPT calls the Nektar++ API to start the simulation. The first task performed inside Nektar++ API is to convert the generic fluid domain mesh generated by the `mesh\_generator' API in python to a mesh that is compatible with Nektar++. Thereafter, MyCrunchGPT calls the Nektar++ API again to initiate the simulation (Figure \ref{fig:prompt8_out}).

\begin{figure}[h]
  \centering
  \centerline{\includegraphics[width = 1\textwidth]{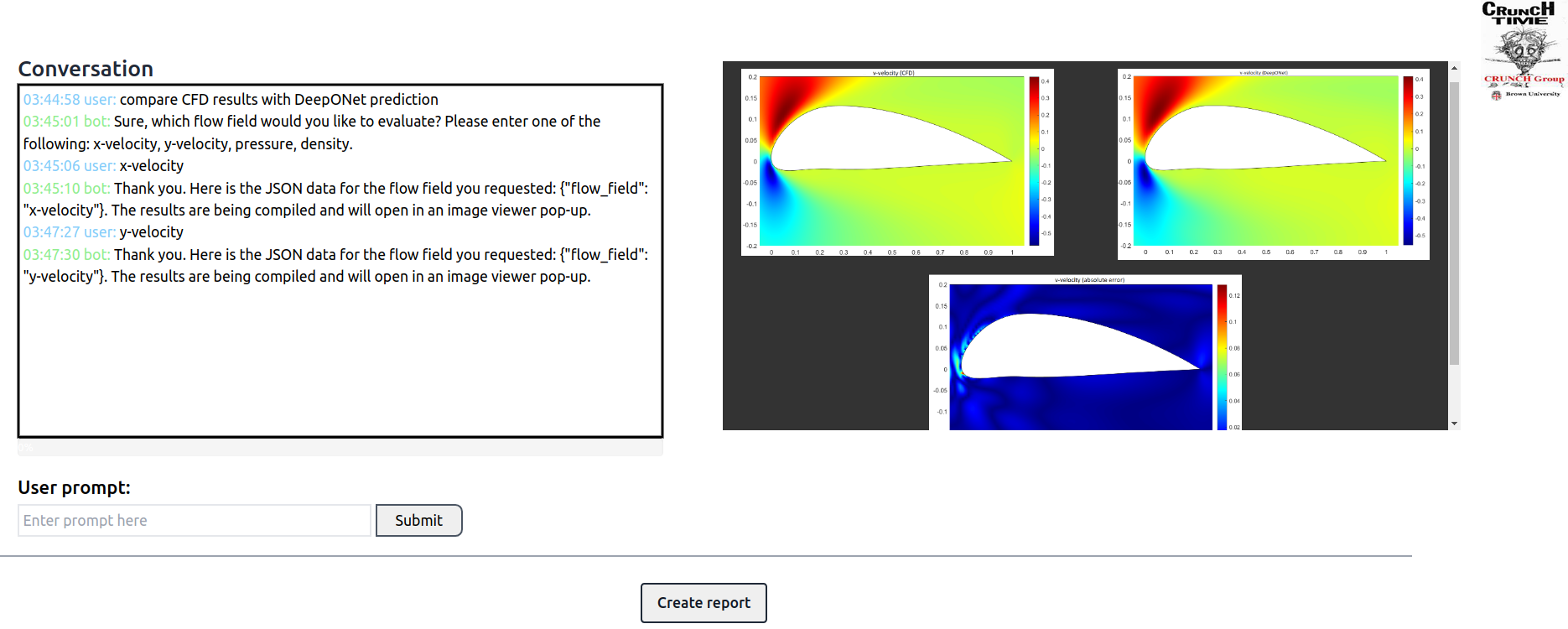}}
  \caption{\textbf{[Request 9]} User queries MyCrunchGPT to show flow field comparisons. Here, the y-velocity comparison between Nektar++ CFD simulation and DeepONet prediction for the optimized geometry is shown in the image viewer.}
  \label{fig:prompt9_out}
\end{figure}

\begin{figure}[h]
     \centering
     \begin{subfigure}[t]{0.33\textwidth} 
         \centering
         \centerline{\includegraphics[width = 1\textwidth]{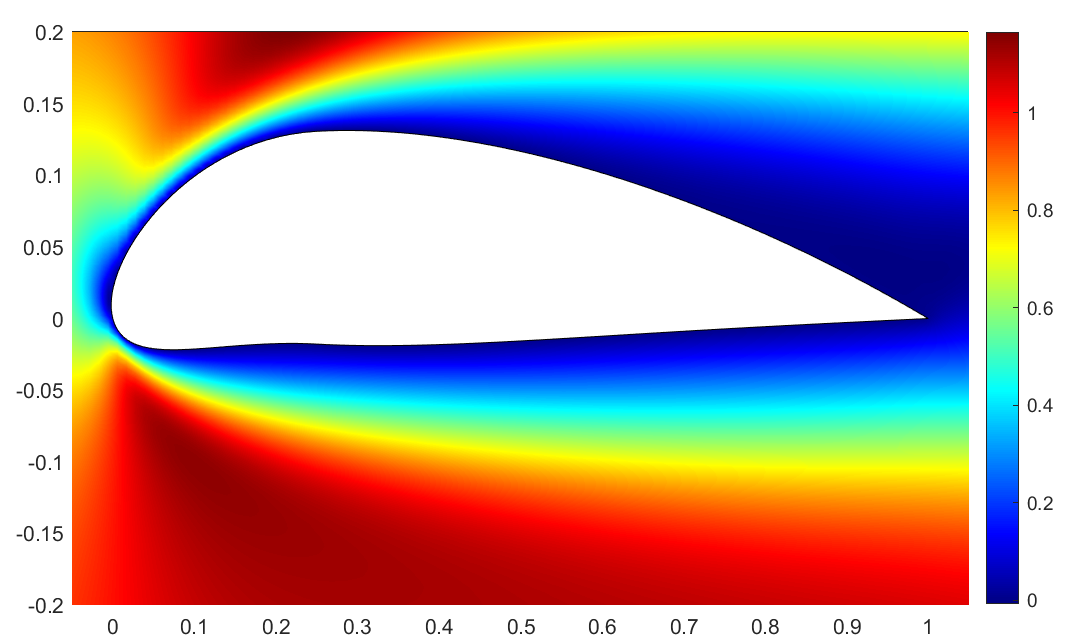}}
         \caption{CFD results for x-velocity}
         \label{fig:CFD_x}
     \end{subfigure}
     \hfill
     \begin{subfigure}[t]{0.325\textwidth} 
         \centering
         \centerline{\includegraphics[width = 1\linewidth]{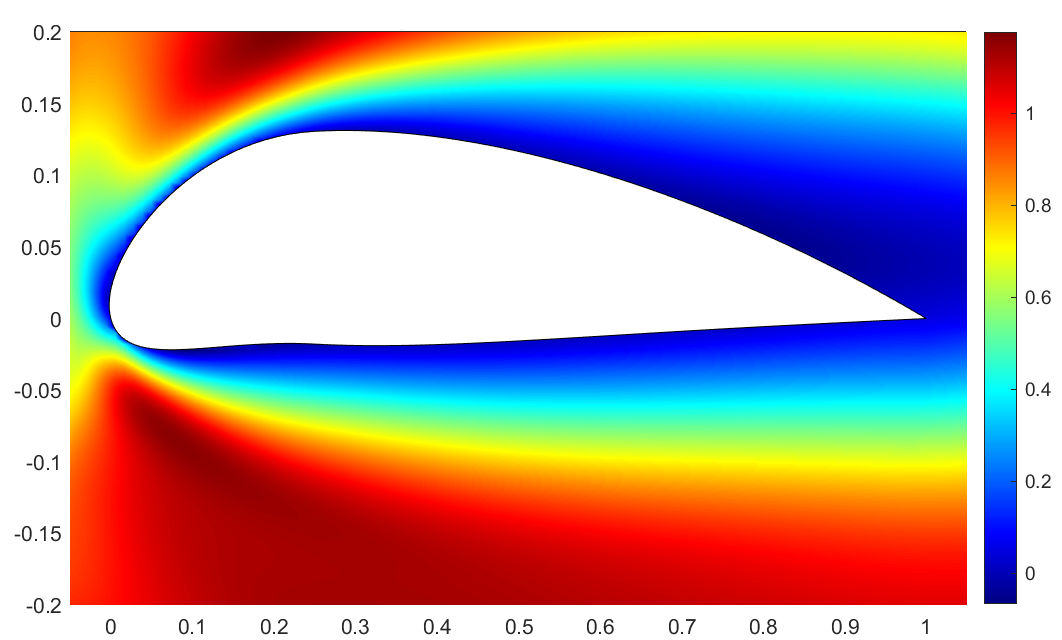}}%
         \caption{DeepONet prediction for x-velocity}
         \label{fig:DON_x}
     \end{subfigure}%
     \hfill
         \begin{subfigure}[t]{0.325\linewidth} 
         \centering
         \centerline{\includegraphics[width = 1\linewidth]{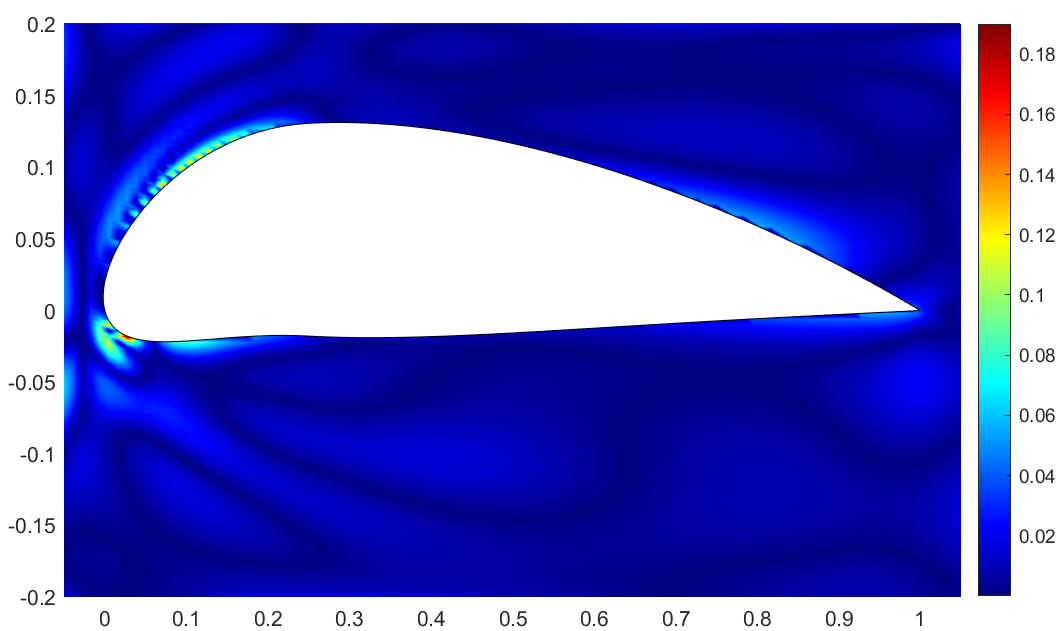}}
         \caption{Absolute error between CFD and DeepONet prediction}
         \label{fig:err_x}
     \end{subfigure}%

    \vfill
     \begin{subfigure}[t]{0.33\textwidth} 
         \centering
         \centerline{\includegraphics[width = 1\textwidth]{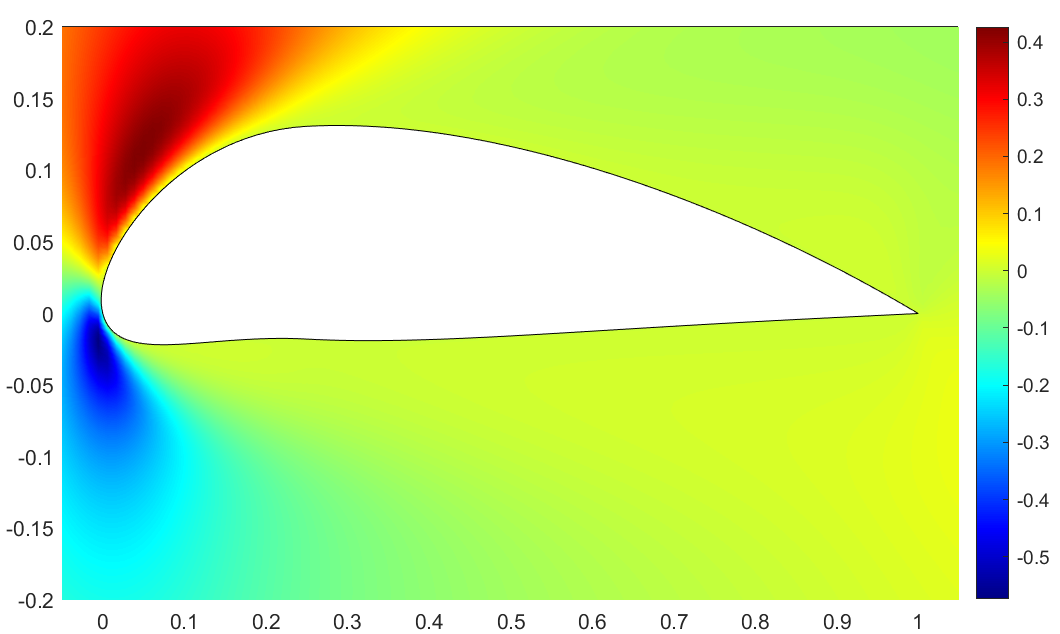}}
         \caption{CFD results for y-velocity}
         \label{fig:CFD_y}
     \end{subfigure}
     \hfill
     \begin{subfigure}[t]{0.325\textwidth} 
         \centering
         \centerline{\includegraphics[width = 1\linewidth]{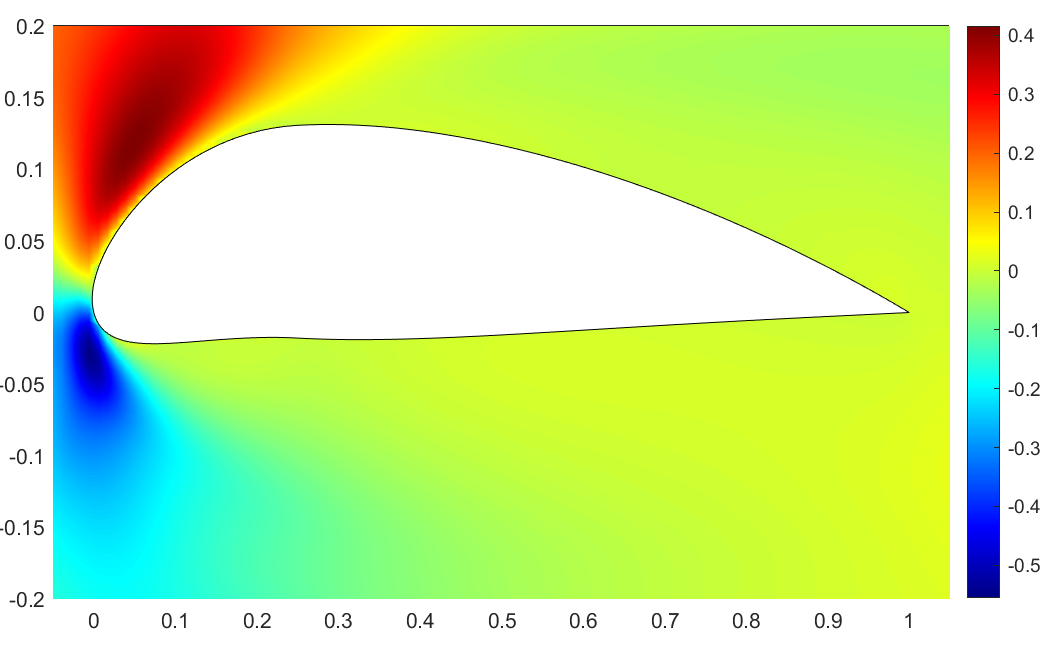}}%
         \caption{DeepONet prediction for y-velocity}
         \label{fig:DON_y}
     \end{subfigure}%
     \hfill
         \begin{subfigure}[t]{0.325\linewidth} 
         \centering
         \centerline{\includegraphics[width = 1\linewidth]{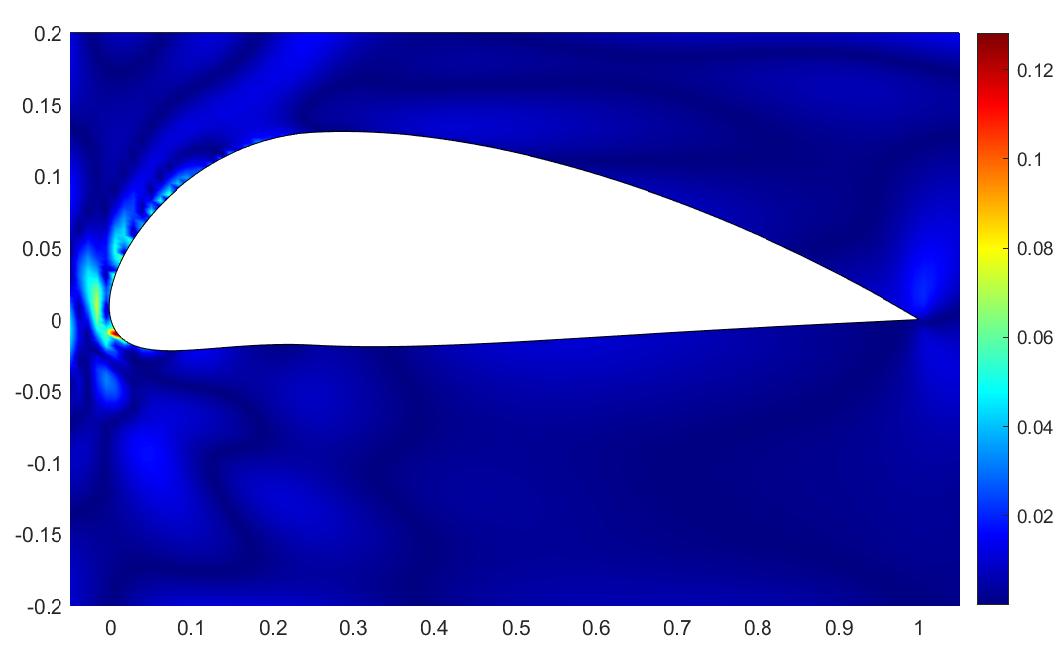}}
         \caption{Absolute error between CFD and DeepONet prediction}
         \label{fig:err_y}
     \end{subfigure}%
     \caption{Comparison for x-velocity between CFD simulation (\ref{fig:CFD_x}) and DeepONet prediction (\ref{fig:DON_x}). When the user requests to see y-velocity comparison, MyCrunchGPT compiles the necessary visualization APIs to generate the comparison: CFD simulation (\ref{fig:CFD_y}), DeepONet predictions (\ref{fig:DON_y}), and their absolute error (\ref{fig:err_y}). These figures here are intended for reader's clarity and evaluation hence plotted outside the image viewer window.}
     \label{fig: vvel_comp}
\end{figure}

\textbf{Request9 -> Compare DeepONet predictions with CFD results:} MyCrunchGPT notifies the user once simulation is complete. Thereafter, the user may request MyCrunchGPT to compare the CFD results with DeepONet prediction for the optimized airfoil geometry. At this stage, MyCrunchGPT requests the user to provide the flow field (x-velocity) that needs to be compared, parses the user input, and generates the necessary comparison. The user can then request for additional flow field comparison (y-velocity) and MyCrunchGPT generates the appropriate comparison using contextual semantics(figure \ref{fig:prompt9_out}). We show the error comparison between CFD and DeepONet in separate plots for reader's clarity (Figure \ref{fig:CFD_x} - \ref{fig:err_x}). (Figure \ref{fig:CFD_y} - \ref{fig:err_y}).


\subsection{MyCrunchGPT assisted result validation and comparison} 

\begin{figure}[h]
  \centering
  \centerline{\includegraphics[width = 0.6\textwidth]{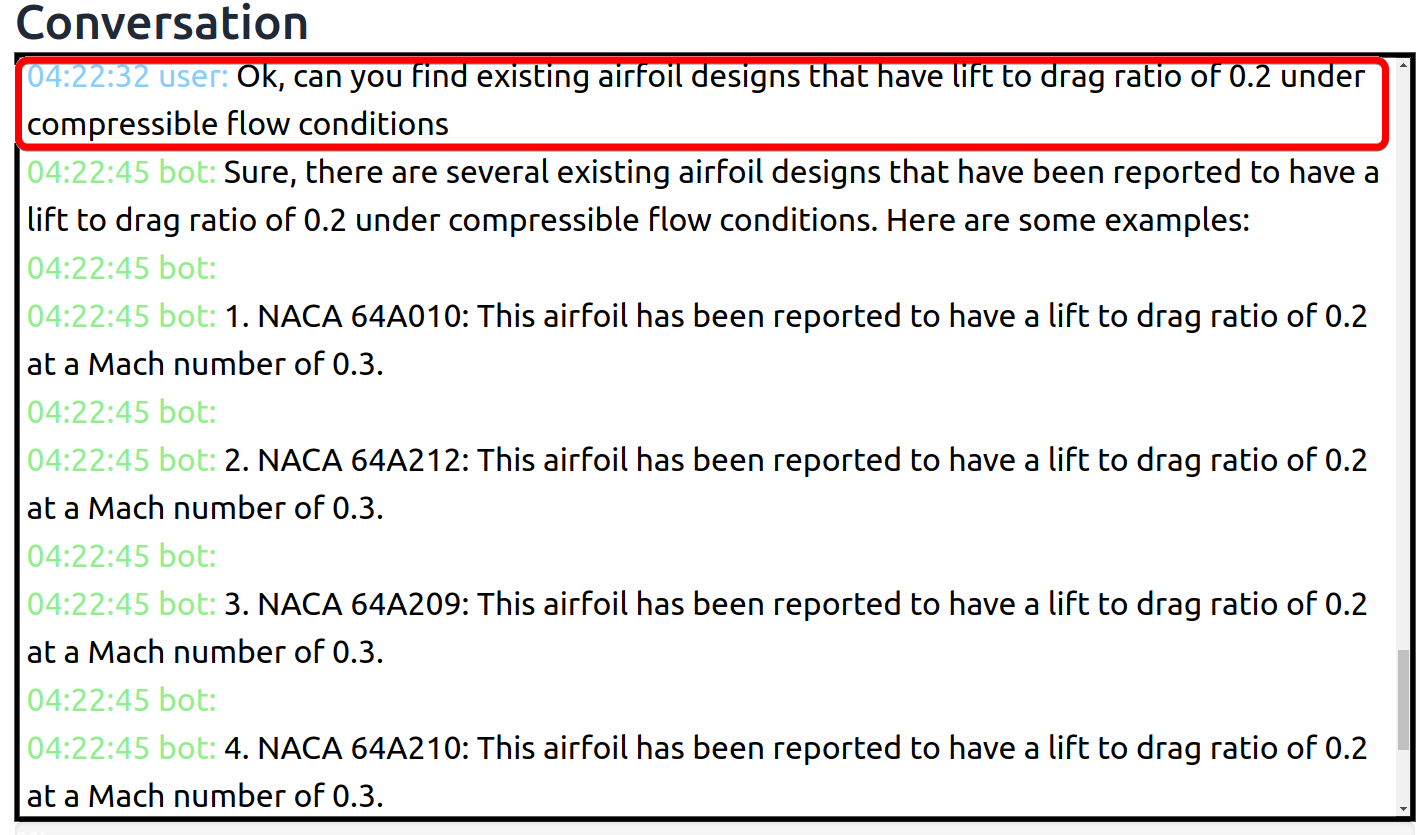}}
  \caption{The user requests MyCrunchGPT to compare the specifications of the optimized airfoil geometry with existing airfoil designs found in literature. MyCrunchGPT searches for relevant airfoil designs with a drag to lift ratio of 0.2 and provides a relevant response that can be investigated further by the user.}
  \label{fig:LD_airfoils}
\end{figure}

The extensive language training data used for default ChatGPT model training can also be used for validating and comparing results with existing literature. Here, we demonstrate a query based result validation process through MyCrunchGPT. We query MyCrunchGPT to identify existing airfoil designs with lift to drag ratio of 0.2, which is similar to was found during the airfoil optimization process. MyCrunchGPT searches for relevant airfoil designs in literature (restricted to the data used for its training) that report the desired lift to drag ratio (see Figure \ref{fig:LD_airfoils}). The user can further query MyCrunchGPT to identify the necessary details regarding a relevant airfoil to obtain succinct answers. We note that this validation process depends on ChatGPT's ability to search for relevant results from within its training data and may not be accurate for all search cases. In future work, we intend to tune MyCrunchGPT's response to improve its search accuracy and response by creating an instruction set to search appropriate literature databases.

\section{Exemplar for AI-assisted PINN-based analysis} 
    As an example of AI-assisted creation of PINN codes, we will ask MyCrunchGPT to solve the lid-driven cavity problem, a standard benchmark problem in CFD for steady-state incompressible flow. The PDE is defined as
    \begin{align}\label{ns_eq}
    \begin{aligned}(\mathbf{u}\cdot\nabla)\mathbf{u} &= -\nabla p + \frac{1}{Re}\nabla^2\mathbf{u}\quad \text{in }\Omega \\
        \nabla\cdot\mathbf{u} &= 0\quad \text{in }\Omega
    \end{aligned}
    \end{align}

    where $\Omega$ is generally taken to be the unit square $[0, 1] \times [0, 1]$. Here $\bm{u}$ and $\bm{p}$ are velocity vector and pressure, respectively. The Dirichlet boundary condition on top boundary of the domain is applied by setting the lid velocity to 1 in the x-direction, and no-slip boundary conditions applied to the other walls by setting both components of the velocity to zero. $Re$ is the Reynolds number, set to 100.

\begin{figure}[h]
     \centering
     \begin{subfigure}[b]{0.45\textwidth}
         \centering
         \centerline{\includegraphics[width = 0.7\textwidth]{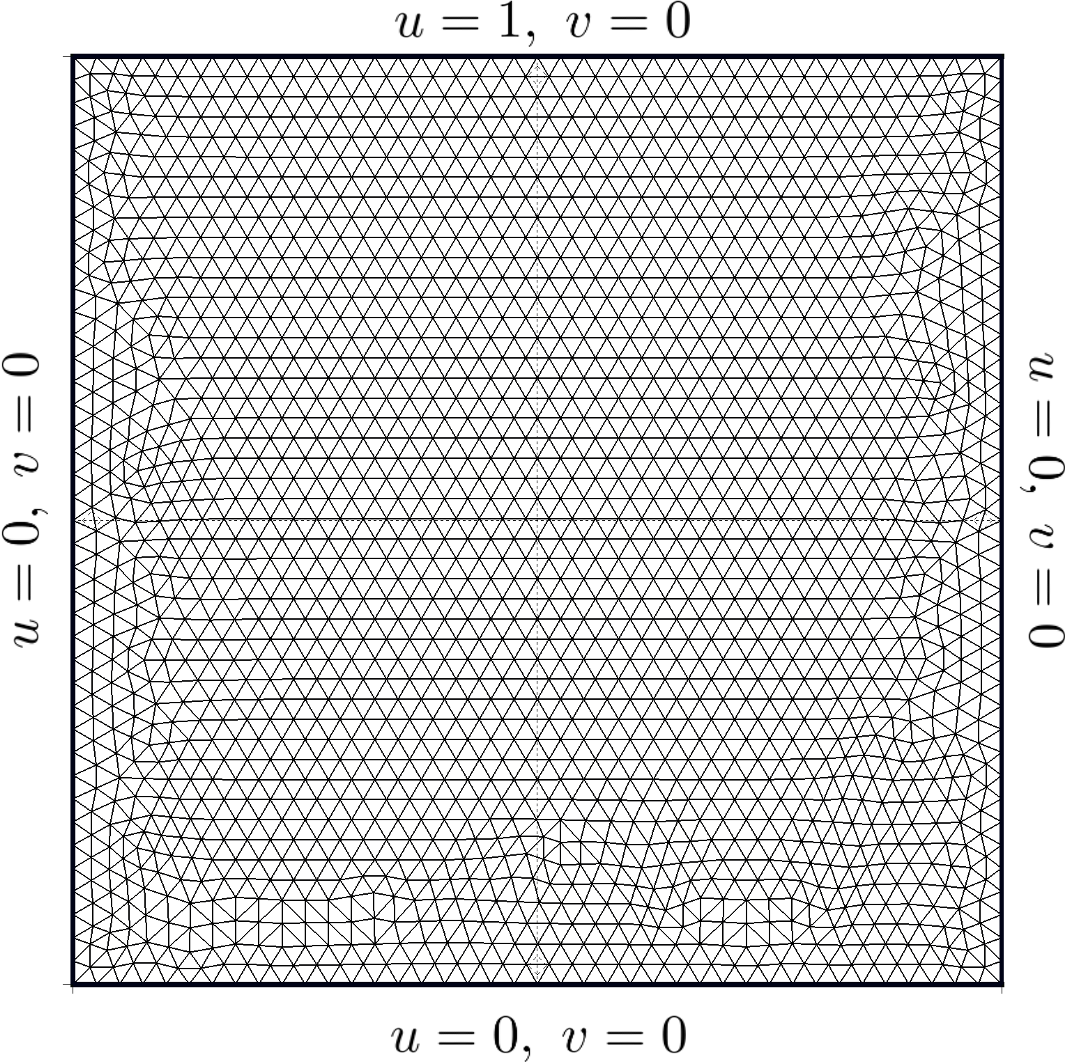}}
         \caption{}
     \end{subfigure}
     \hfill
 \begin{subfigure}[b]{0.45\textwidth}
         \centering
         \centerline{\includegraphics[width = 0.7\textwidth]{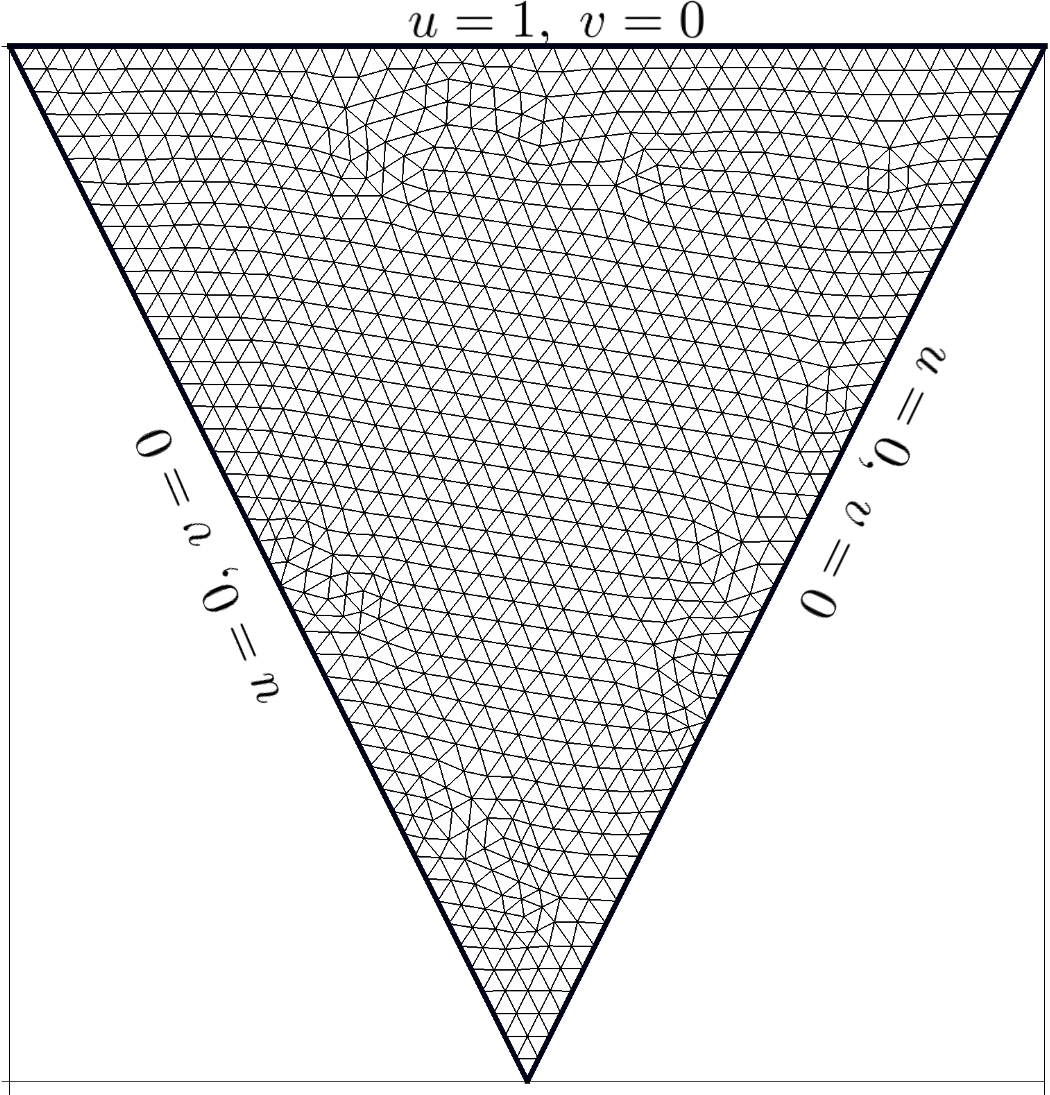}}
         \caption{} 
     \end{subfigure}

     \caption{Discretization of (a) square (b) triangular domain  by using unstructured triangular elements used for solving the equation \eqref{ns_eq} to validate the velocity fields obtained from PINNs. Equation \eqref{ns_eq} is solved by spectral element method provided by Nektar++ frameworks. The boundaries of the domain are marked by respective boundary condition used in formulating the spectral element method. Here $u$ and $v$ represent the horizontal and vertical velocity of the fluid, respectively.}
     \label{fig:domain_discrete}
\end{figure}

\begin{figure}[h!]
    \centering
    \includegraphics[width=0.9\textwidth]{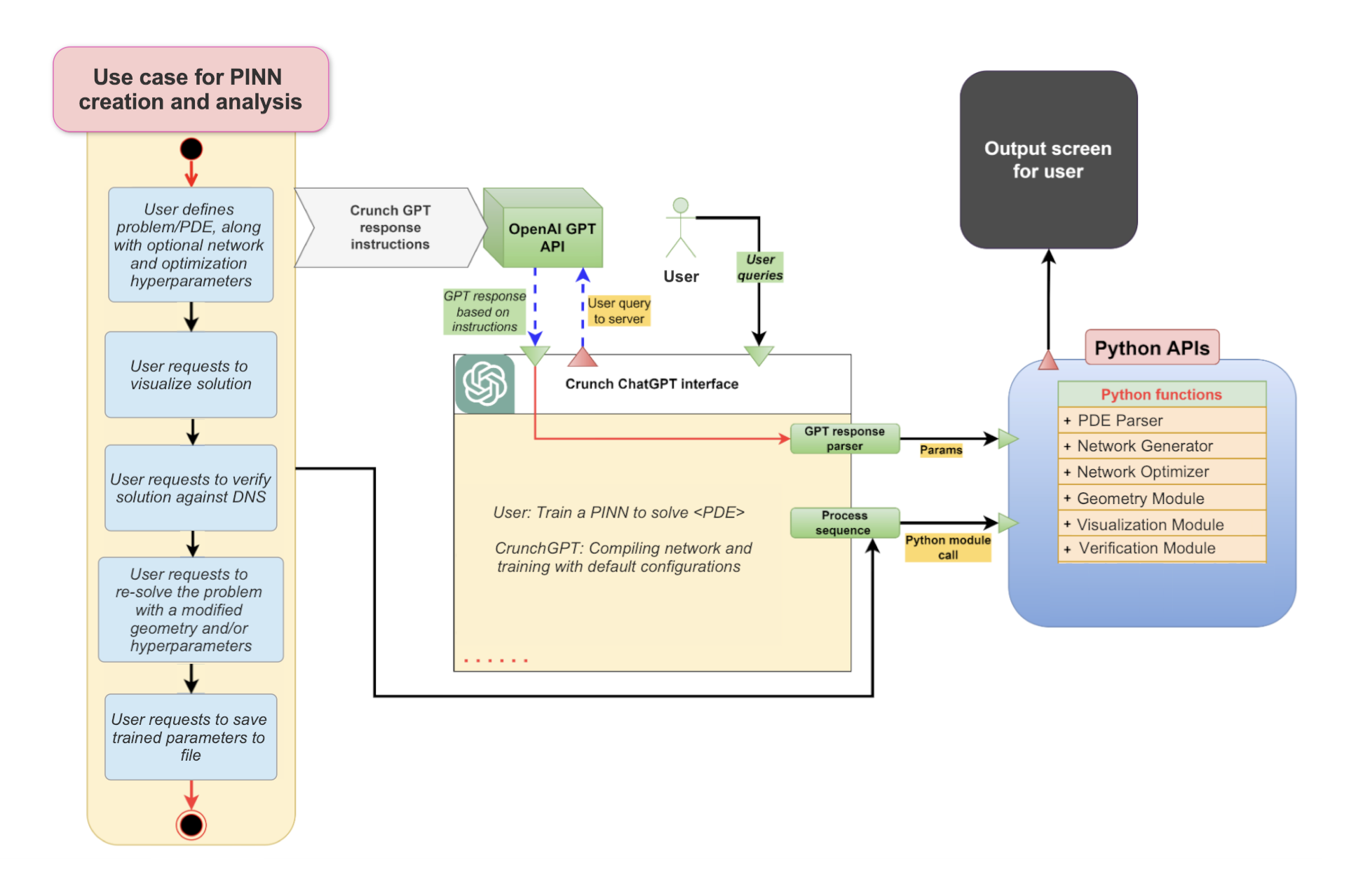}
    \caption{Workflow diagram for asking MyCrunchGPT to solve a PDE with a PINN. The user defines the PDE and relevant details, and asks to train a PINN for the PDE. The user may provide various requests, such as specifying configurations, visualizing different components of the solution, verifying against known solutions/DNS, or saving/loading trained parameters. The user may ask to modify the problem, referring to the previous instructions without explicitly re-defining the PDE. GPT is used to parse the instructions into components, map those components to the corresponding internal modules, and synthesize the full program by composing the selected modules.}
    \label{fig:pinn_workflow}
\end{figure}

\begin{figure}[h!]
    \centering
    \includegraphics[width=1\textwidth]{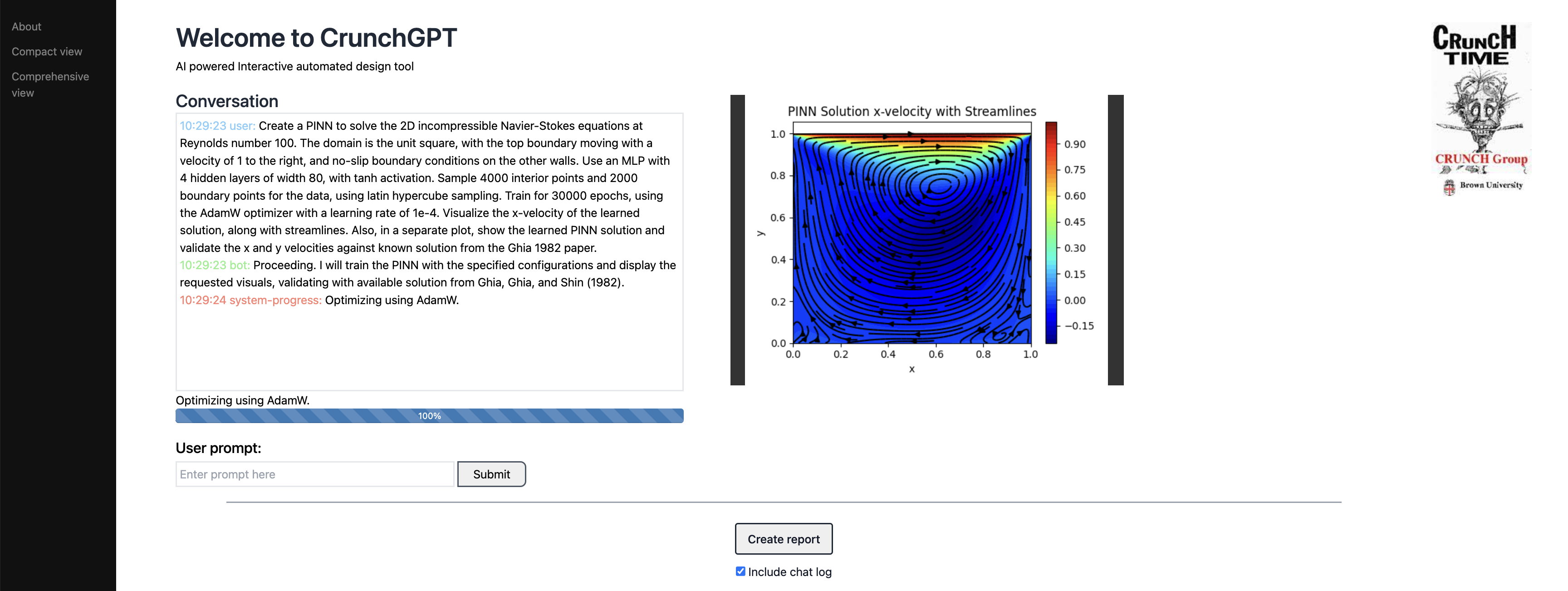}
    \caption{Asking MyCrunchGPT to use a PINN to solve the lid-driven cavity CFD benchmark problem. The user specifies the PDE to solve, along with the geometry and boundary conditions in natural language, as well as training configurations. The user also requests to visualize the solution in two different ways, one involving validation against a specific known solution. MyCrunchGPT parses the request into the necessary components and synthesizes and executes a program to complete the request. The image in the image view window }
    \label{fig:pinn_chat_11}
\end{figure}

\begin{figure}[h!]
    \centering
    \includegraphics[width=1\textwidth]{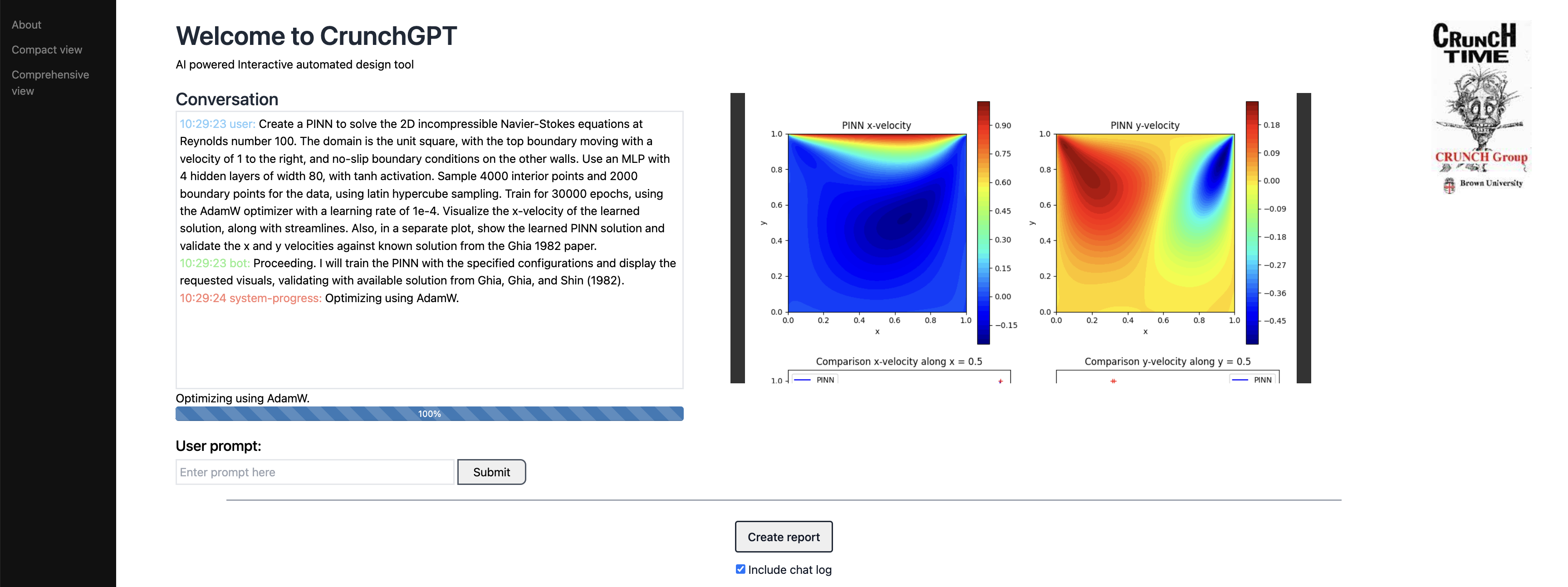}
    \caption{MyCrunchGPT interface with the first half of the second requested visualization (visualizing the PINN solution).}
    \label{fig:pinn_chat_12}
\end{figure}

\begin{figure}[h!]
    \centering
    \includegraphics[width=1\textwidth]{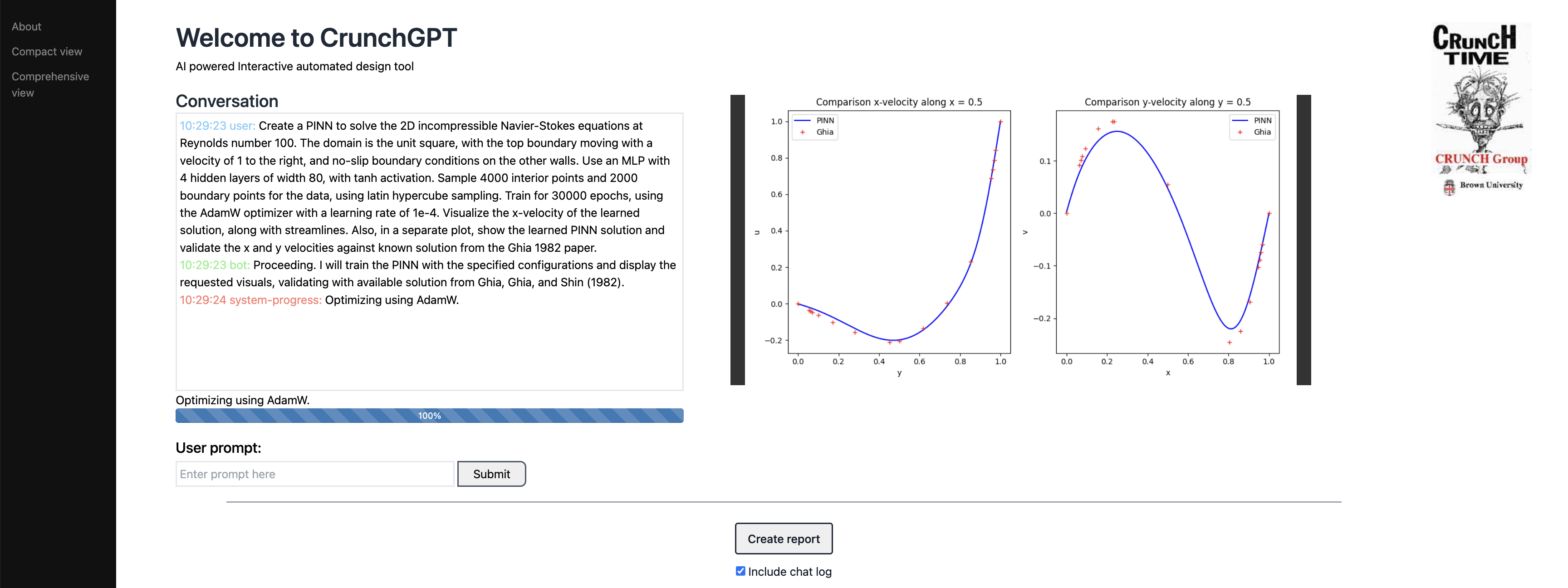}
    \caption{MyCrunchGPT interface with the second half of the second requested visualization (validation with specified known solution from \cite{Ghia1982HighReSF}).}
    \label{fig:pinn_chat_13}
\end{figure}

\begin{figure}[h!]
    \centering
    \includegraphics[width=1\textwidth]{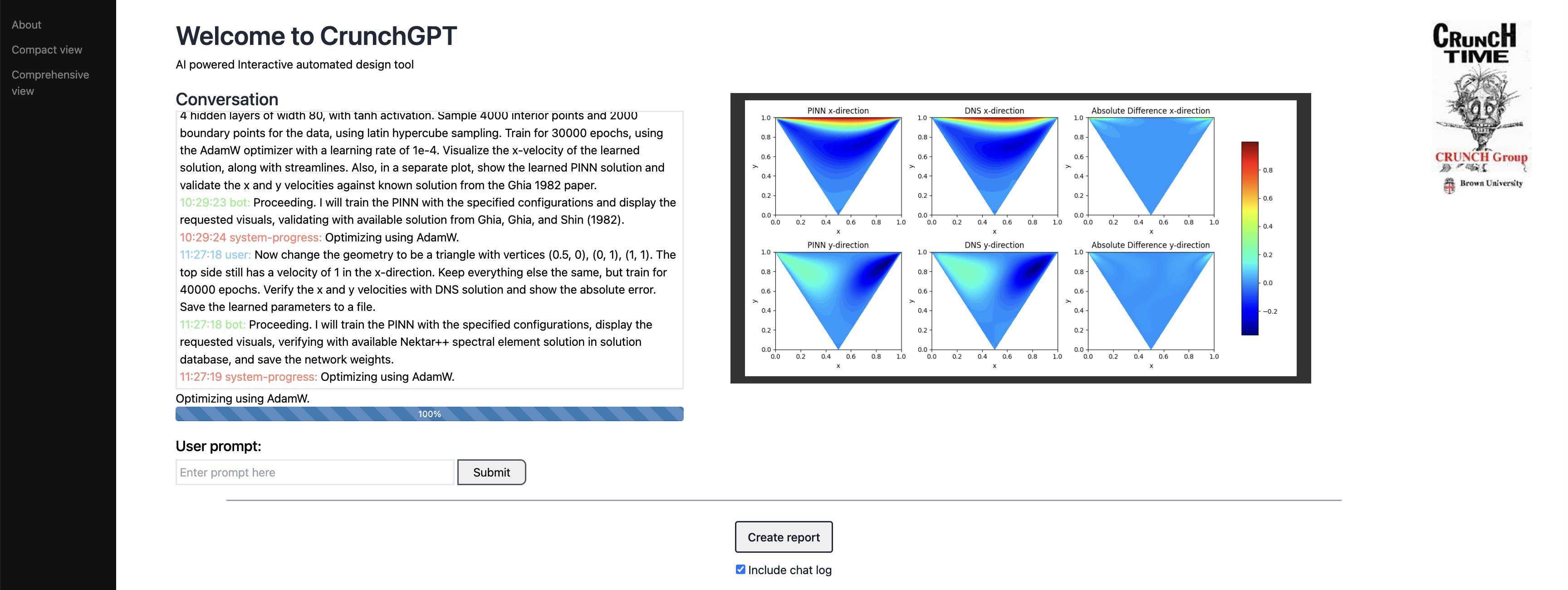}
    \caption{Asking MyCrunchGPT to solve a variation of the previous problem. The user requests to change the geometry and training iterations, verify with any DNS solution available, and save the learned parameters, for which MyCrunchGPT modifies and executes the code appropriately.}
    \label{fig:pinn_chat_21}
\end{figure}

In this work, we present the ChatGPT assisted PINNs for lid driven cavity flow problem on triangular and square domains. We chose two different types of geometry and discretized it with unstructured triangular meshes, as shown in Figure \ref{fig:domain_discrete}. The workflow diagram for this problem is given in Figure \ref{fig:pinn_workflow}. Figures \ref{fig:pinn_chat_11}, \ref{fig:pinn_chat_12}, and \ref{fig:pinn_chat_13} show the first set of chat instructions along with the requested visuals. Figure \ref{fig:pinn_chat_21} shows the second set of chat instructions, modifying the problem, along with the requested visuals. To verify and validate the results obtained from MyCrunchGPT, we prepared databases for lid driven cavity flow in square domain by digitizing the results from existing literature \cite{Ghia1982HighReSF} and integrated in the ChatGPT framework. Therefore, results generated from MyCrunchGPT framework comes with a certificate of fidelity.

Internally, MyCrunchGPT has access to various optimized and thoroughly-tested Python codes we developed for geometry (both 2D and 3D), differentiation, training, verification, and visualization. These codes are developed to be highly modular, so that MyCrunchGPT can compose them easily to achieve the user's specifications, allowing for flexibility in user request, while maintaining code correctness and largely avoiding hallucinations that LLMs are prone to, especially if asking to code from scratch. We also have a database of example scientific codes and high-accuracy solutions for many benchmark problems that MyCrunchGPT has access to. MyCrunchGPT for PINN simulation is being continuously developed, with new features such as web search, paper Q\&A, uncertainty quantification, hyperparameter tuning, among others, being added soon. Multimodality will also be added, with the ability to reason interactively with a given solution visual.

\section{Summary}
In this work, we use two exemplars to demonstrate the use of ChatGPT for scientific machine learning (called MyCrunchGPT) and integration with user-defined workflow as an assistant. As a first example, we tailor ChatGPT's response to assist a user with different steps involved in design and optimization of 2D NACA airfoils. For this, we create a set of instructions for MyCrunchGPT to follow based on user's request and a user defined design workflow. The process starts with the user specifying the need to generate new airfoil designs, which are constrained by the parameters maximum camber ($m$) and position of maximum camber ($p$). MyCrunchGPT then prompts the user to provide necessary parameter values through a conversation window, parses user's requests, and calls necessary python functions in the back-end. Once designs are generated, MyCrunchGPT guides the user through the evaluation process through a surrogate DeepONet model. This surrogate model is trained offline with CFD data of airfoil designs, as outlined in \cite{shukla2023deep}, and is used to determine the flow field around these new airfoil designs. As the next step, the user requests MyCrunchGPT to optimize the airfoil shape using constraints on parameters ($m$) and ($p$), with the the objective being to maximize the lift to drag ratio. MyCrunchGPT invokes the necessary optimization routine in the back-end and an optimized airfoil geometry is generated. Finally, MyCrunchGPT also assists the user to compare the DeepONet predicted flow field results with a traditional numerical solver (Nektar++), which is integrated with the MyCrunchGPT workflow. The comparison between DeepONet predictions vs CFD solver prediction for the optimized airfoil geometry show small absolute error. Lastly, MyCrunchGPT provides users with the ability to validate their findings with results available in reported literature. In summary, we demonstrate an instruction-based framework for ChatGPT to invoke appropriate response to guide the user through a design and optimization process.  

As a second example, we use MyCrunchGPT to assist in the creation and training of a PINN to solve a specified PDE (the Navier-Stokes equations for the lid-driven cavity benchmark problem), as well as provide visualization and verification of the learned solutions against data from direct numerical simulation. Further requests pertain to validation of the learned solution against available experimental data in the literature. The PINN capabilities of MyCrunchGPT (and the capabilities of MyCrunchGPT in general) are being actively developed to include useful tools such as uncertainty quantification, multimodal solution interaction, hyperparameter tuning, and distributed computation, among others.

The entire process has been embedded in a GUI, based on a webapp. The webapp allows the user to interact with the bot, view the system logs, and observe the figures and outputs of the process. We employed various ways for the user to interact with the GUI, and plan to extend it so more complicated asks are accommodated by the GUI. For example, if the user asks for a table of comparison of certain results. We intend to add more visualization techniques, and improve the way the front-end interacts with the back-end. Last, when the information that appears on the screen is acceptable by the user, they can export it into a summary document. This document also includes a bottom line summary of the process and comparison to benchmarks, done by the bot. We plan to expand this as well, and allow the user to interact with the bot to create more tailored reports, depending on the application.

In future work on MyCrunchGPT, we plan to develop probabilistic set based design \cite{ward1995second_sbd, sobek1999_sbd, royset2017_sbd, toche2020_sbd, bonfiglio2019_sbd, shintani2022_sbd}, incorporate multiple neural operators such as FNO, WNO, LNO and ViTO \cite{wen2022u_fno, li2020_fno, tripura2023wavelet_wno, cao2023_lno, ovadia2023vito}, and include different versions of PINNs, including domain decomposition and adaptive activation functions \cite{xpinn_shukla2021, jagtap2020adaptive, qian2018adaptive}, that will make CruchGPT more versatile and robust. We will also develop new versions of MyCrunchGPT appropriate for other areas, including systems biology,  solid mechanics, materials science, geophysics, and bioinformatics. 

\section*{Acknowledgements}
This work was supported by the DOE SEA-CROGS project (DE-SC0023191), the MURI-AFOSR FA9550-20-1-0358 project and by the ONR Vannevar Bush Faculty Fellowship (N00014-22-1-2795).

\newpage
\bibliographystyle{unsrt}  
\bibliography{references}

\end{document}